\documentclass[runningheads]{llncs}

\usepackage{graphicx}
\usepackage{booktabs}
\usepackage{pifont}
\usepackage{xcolor}
\usepackage{arydshln}
\usepackage[accsupp]{axessibility}  

\usepackage{enumitem} 
\usepackage{listings} 
\usepackage{xcolor} 

\lstdefinestyle{jsonstyle}{  
  basicstyle=\ttfamily\small,
  breaklines=true,
  breakatwhitespace=false,
  columns=fullflexible,
  keepspaces=true,
  showstringspaces=false,
  frame=single,
  framerule=0.3pt
}

\usepackage{fvextra} 
\fvset{
  breaklines=true,
  breakanywhere=true,
  breaksymbol={},
  breaksymbolleft={},
  breaksymbolright={}
}

\usepackage{listings}
\lstdefinestyle{promptstyle}{
  basicstyle=\small\normalfont,
  columns=fullflexible,
  breaklines=true,
  breakatwhitespace=false,
  keepspaces=true,
  showstringspaces=false,
  breakindent=0pt,
  breakautoindent=false,
  prebreak=\mbox{},
  postbreak=\mbox{},
  literate=
    {-- }{{$\bullet$\ }}2
    {°}{{$^\circ$}}1
    {“}{{``}}1
    {”}{{''}}1
    {’}{{'}}1
    {–}{{--}}1
}

\usepackage[T1]{fontenc}
\usepackage{textcomp}

\usepackage[pagebackref,breaklinks,colorlinks]{hyperref}

\usepackage{orcidlink}
\usepackage{amsmath}
\usepackage{amssymb}
\usepackage{mathtools}
\usepackage[table,svgnames,dvipsnames]{xcolor}
\usepackage[most]{tcolorbox}
\usepackage{microtype}
\usepackage{subcaption}
\usepackage{multirow}
\usepackage{wrapfig}
\usepackage{placeins}
\usepackage{needspace}
\usepackage{makecell}
\usepackage{caption}
\usepackage[capitalize,noabbrev]{cleveref}
\newcommand{\name}[1]{RS-WorldModel}
\newcommand{\data}[1]{RSWBench-1.1M}
\newcommand\blfootnote[1]{
  \begingroup
  \renewcommand\thefootnote{}\footnote{#1}%
  \addtocounter{footnote}{-1}%
  \endgroup
}
\usepackage{multibib}
\begin{document}

\title{\raisebox{-0.2em}{\includegraphics[height=1.2em]{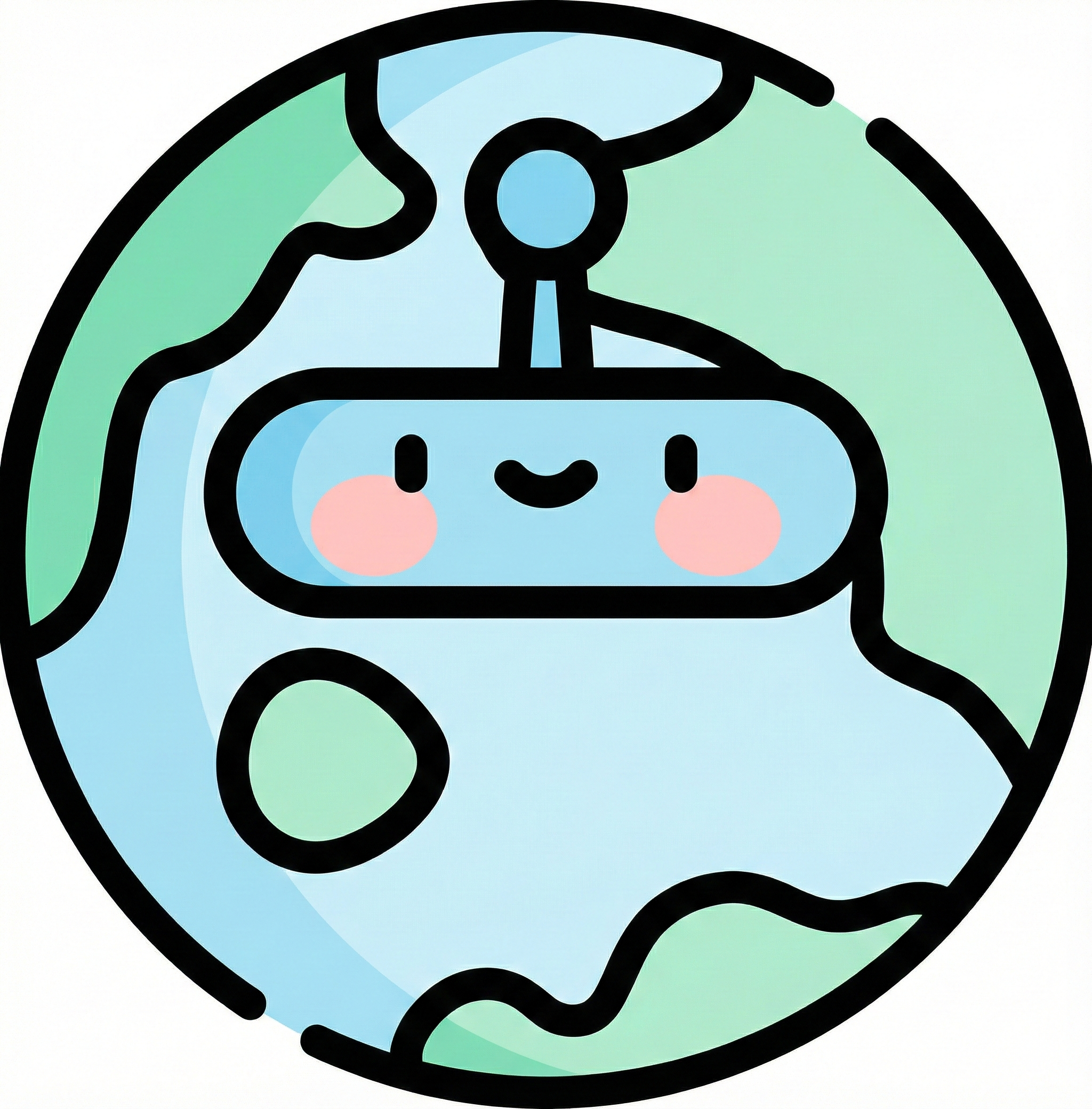}}~\name{}: a Unified Model for Remote Sensing Understanding and Future Sense Forecasting} 

\titlerunning{\name{}}




\author{Linrui Xu\inst{1}$^*$ \and
Zhongan Wang\inst{2}$^*$ \and
Fei Shen\inst{3} \and
Gang Xu\inst{4} \and
Huiping Zhuang\inst{5} \and \\
Ming Li\inst{4}$^\dagger$ \and
Haifeng Li\inst{1}$^\dagger$}

\authorrunning{L.~Xu et al.}

\institute{
\makebox[0.9\textwidth][c]{\inst{1}Central South University, \inst{2}Zhejiang University, \inst{3}National University of Singapore} \\
 \inst{4}Guangming Laboratory, 
 \inst{5 }South China University of Technology
}

\maketitle

\blfootnote{
\noindent
$^*$ Equal contribution. \\
$^\dagger$ Corresponding author.(\email{ming.li@u.nus.edu, lihaifeng@csu.edu.cn}) \\
$^1$ Codes and datasets are available at https://github.com/GeoX-Lab/RS-WorldModel
}

\begin{abstract}

Remote sensing world models aim to both explain observed changes and forecast plausible futures, two tasks that share spatiotemporal priors. Existing methods, however, typically address them separately, limiting cross-task transfer. We present \name{}, a unified world model for remote sensing that jointly handles spatiotemporal change understanding and text-guided future scene forecasting, and we build \data{}, a \textbf{1.1 million}  sample dataset with rich language annotations covering both tasks. \name{} is trained in three stages: (1) Geo-Aware Generative Pre-training (GAGP) conditions forecasting on geographic and acquisition metadata; (2) synergistic instruction tuning (SIT) jointly trains understanding and forecasting; (3) verifiable reinforcement optimization (VRO) refines outputs with
verifiable, task-specific rewards. With only 2B parameters, \name{} surpasses open-source models up to 120$  \times  $ larger on most spatiotemporal change question-answering metrics. It achieves an FID of 43.13 on text-guided future scene forecasting, outperforming all open-source baselines as well as the closed-source Gemini-2.5-Flash Image (Nano Banana)\textsuperscript{{1}}.
  \keywords{Cross-modal understanding and generation \and World Model \and Remote Sensing}


\end{abstract}
\section{Introduction}
\label{sec:intro}

\begin{figure}[t]
  \centering
\includegraphics[width=\linewidth]{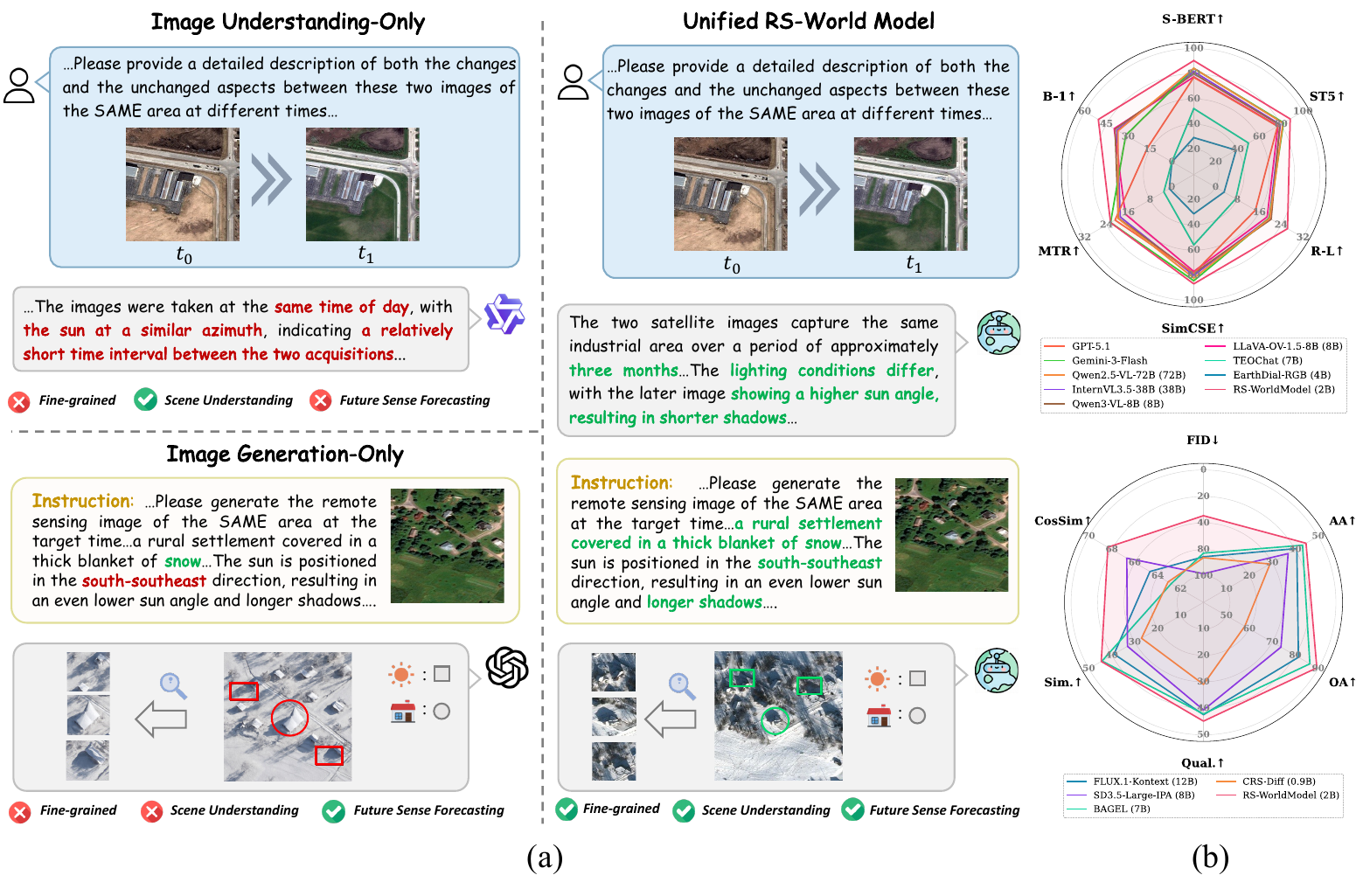}
\caption{\name{}: A unified world model for remote sensing that integrates spatiotemporal change understanding and future scene forecasting capabilities. (a) Qualitative comparison with leading methods. (b) Quantitative results on the \data{}.}
  \label{fig:teaser}
\vspace{-0.3em}
\end{figure}

World models, which construct internal representations of environments and predict their future dynamics, have become an active research direction in application domains such as autonomous driving, robotics, and generative simulation~\cite{ding2025understanding}.
In autonomous driving, GAIA-1~\cite{hu2023gaia} and Drive-WM~\cite{wang2024driving} forecast driving scenes conditioned on planned actions and map context.
Video generation systems such as Sora~\cite{brooks2024video} demonstrate that large-scale generative models can serve as versatile physical simulators.
In embodied AI, DayDreamer~\cite{wu2023daydreamer} trains robot locomotion and manipulation policies primarily within a learned world model, while Cosmos~\cite{agarwal2025cosmos} proposes a general-purpose world foundation model trained on massive video data.
These efforts converge on a shared insight: learning to predict future states encourages a model to internalize environment dynamics, making world models a promising path toward general purpose autonomous agents.
Earth observation, where satellites repeatedly image the same locations over time, stands to benefit substantially, yet remains unexplored (\cref{fig:teaser}).

Recent remote sensing generative models~\cite{zhao2025change,khanna2023diff} can synthesize plausible satellite imagery, but they are typically confined to pixel-level synthesis without reasoning about \emph{what} changed or \emph{why}.
Conversely, understanding-oriented models~\cite{kuckreja2024geochat,hu2025rsgpt,zhan2025skyeyegpt} interpret observed scenes but are not designed for future or counterfactual states.
In many remote sensing settings, applications need \emph{both} accurate interpretation and controllable forecasting~\cite{bastani2023,van2021multi,mall2023change}. Both tasks depend on shared priors from geographic and acquisition context (e.g., location, seasonality, and sensor characteristics). Training them separately fails to exploit this shared structure, leaving generation difficult to control and understanding unable to leverage dense generative supervision~\cite{zeng2025futuresightdrive,koksal2025few}.

Building a unified remote sensing world model poses three core challenges.
\textit{First}, to the best of our knowledge, no existing dataset simultaneously supports spatiotemporal change understanding and future scene forecasting at scale; most benchmarks~\cite{chen2025rscc,revankar2025,christie2018} target a single task and lack the rich geographic metadata needed for location-aware modeling.
\textit{Second}, remote sensing imagery exhibits complex spatiotemporal variations driven by geographic location, sensor parameters, and seasonal cycles, making it difficult to learn effective generation priors from limited data~\cite{xuan2025dynamicvl,wang2025disasterm3,cong2022satmae}. Existing approaches train understanding and generation in isolation~\cite{zhu2025skysense,muhtar2024lhrs}, limiting knowledge transfer between the two.
\textit{Third}, standard reinforcement learning from human feedback relies on learned preference models that fail to capture the geographic consistency and physical plausibility constraints specific to remote sensing~\cite{koksal2025few}.

We address these challenges with \name{} and \data{}.
For data, we construct \data{}, a large-scale dataset of \textbf{1.1M} high-resolution samples covering both spatiotemporal change understanding and text-guided future scene forecasting, enriched with fine-grained geographic metadata and built on fMoW~\cite{christie2018} to ensure global diversity.
For modeling, we propose \name{}, the first unified world model for remote sensing, trained in three stages:
(1)~Geo-Aware Generative Pre-training (GAGP) injects geographic conditioning to establish spatiotemporal forecasting priors;
(2)~synergistic instruction tuning (SIT) jointly optimizes understanding and generation to improve controllability and let each task reinforce the other;
and (3)~verifiable reinforcement optimization (VRO) improves robustness by refining outputs with task-specific verifiable rewards instead of a learned preference model.
With only \textbf{2B} parameters, \name{} surpasses open-source models up to 120$\times$ larger on most spatiotemporal change QA metrics and achieves an FID of \textbf{43.13} on text-guided future scene forecasting, outperforming all open-source baselines and the closed-source Gemini-2.5-Flash Image on FID.
Our contributions are as follows:
\begin{itemize}
    \item We propose \name{}, the first unified world model for remote sensing that jointly handles spatiotemporal change understanding and text-guided future scene forecasting.
    \item We construct \data{}, a large-scale dataset of 1.1M samples covering both tasks with rich geographic metadata and fine-grained language annotations.
    \item We design a three-stage training paradigm (GAGP, SIT, and VRO) that enables a 2B parameter model to outperform far larger open-source models and several closed-source models.
\end{itemize}

\section{RSWBench-1.1M Dataset}
\label{sec:dataset}

\begin{figure*}[t]
  \centering
  \includegraphics[width=\textwidth]{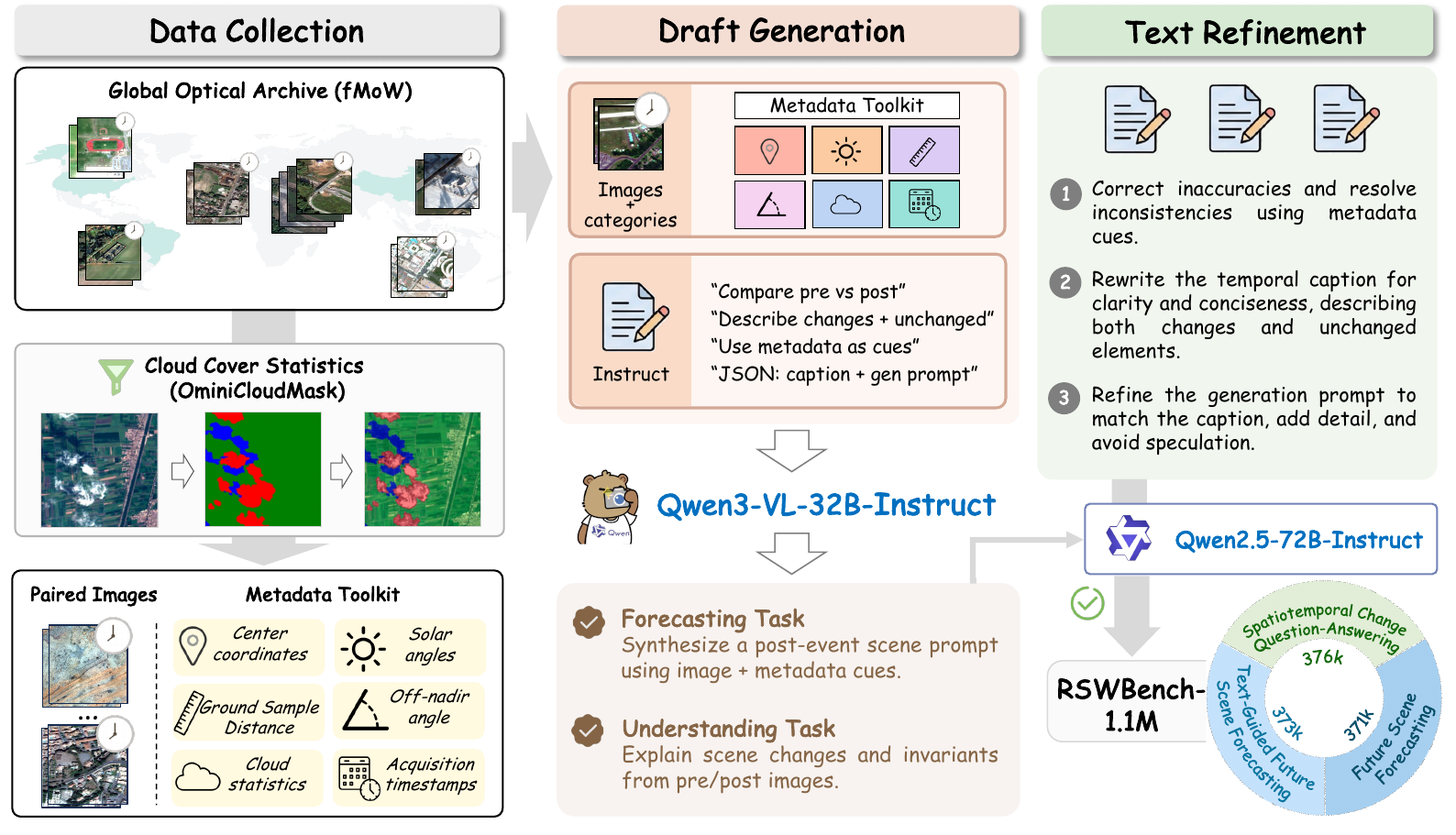}
\caption{\textbf{Data construction pipeline and dataset composition.} We establish a scalable pipeline to transform multi-temporal observations into high-quality instruction data, supporting understanding and forecasting tasks with strict train-test isolation.}
  \label{fig:data_collection}  
\vspace{-0.3em}
\end{figure*}

Training a unified remote sensing world model requires data supporting two core capabilities: \textit{Spatiotemporal Change Question-Answering} (ST-CQA) and \textit{Text-Guided Future Scene Forecasting} (TFSF). We contribute a scalable automated annotation pipeline and a dataset suite with a 1.1M training corpus and a 5.6K evaluation benchmark. Both are derived from the fMoW archive, with strict adherence to official split protocols to prevent data leakage (\cref{fig:data_collection}).
 
\subsection{Scalable Data Construction Pipeline}
Constructing a million-scale dataset with spatiotemporal consistency requires overcoming two challenges: atmospheric noise and the lack of dense semantic annotations. We address these via a two-stage pipeline that unifies physical filtering with semantic refinement.

\noindent\textbf{Stage 1: Physical Standardization.} We first pair multi-temporal observations from the same geographic coordinates. To ensure the model learns from valid ground features rather than artifacts, we normalize acquisition metadata (e.g., sun angles) and filter samples based on visibility. Using OmniCloudMask~\cite{wright2025training}, we estimate the pixel-wise cloud ratio $\rho_{\text{Cloud}}(I)$ and retain only samples where $\rho_{\text{Cloud}}(I) \le 0.9$, discarding only near-total occlusions. Unlike conventional remote sensing datasets that enforce strict clear-sky filters (e.g.\ $\le$5--10\%)~\cite{almar2025global}, we deliberately retain partially cloudy scenes because cloud cover serves as a controllable condition for text-guided Forecasting.

\noindent\textbf{Stage 2: Semantic Refinement.} To synthesize high-quality language supervision without expensive manual annotation, we employ a generate-and-refine strategy. A vision-language model first drafts structured JSON annotations based on image pairs and metadata. Subsequently, a larger, more capable model (Qwen2.5-72B-Instruct) refines these drafts. A key design choice is metadata translation: the pipeline explicitly converts raw numeric sensor data into natural linguistic cues (e.g., translating solar elevation into shadow descriptions), preventing the model from overfitting to numerical values.

\subsection{RSWBench-1.1M Dataset Suite}
Using the pipeline described above, we curate two distinct subsets to support the training and evaluation of remote sensing world models (\cref{tab:rswbench_comparison}).

\noindent\textbf{Training.} Constructed exclusively from the fMoW training split, this corpus contains approximately 1.1M samples. It includes 371K instances for generative pre-training and 742K mixed instances for synergistic instruction tuning. An additional 16K subset is reserved for reinforcement alignment. 


\noindent\textbf{Evaluation.} To establish a rigorous standard, we curate 6.6K samples exclusively from the fMoW test split. The benchmark is balanced, containing 5K ST-CQA and 1.6K TFSF samples. By preserving the global diversity of the original test set, RSWBench-1.1M enables stable evaluation of cross-region generalization and forecasting fidelity.


\begin{table}[t]
\centering
\caption{Comparison of dataset capabilities and scales for remote sensing understanding and generation. \checkmark: supported; \textcolor{gray}{\ding{55}}: not supported; --: not applicable (dataset does not target this task category).}
\label{tab:rswbench_comparison}
\footnotesize
\resizebox{\linewidth}{!}{%
\begin{tabular}{lcccccc}
\toprule
\multirow{2}{*}{Dataset}
& \multirow{2}{*}{\textbf{Scale}}
& \multicolumn{2}{c}{\textbf{Understanding}}
& \multicolumn{3}{c}{\textbf{Generation}} \\
\cmidrule(lr){3-4} \cmidrule(lr){5-7}
& 
& Temporal
& \makecell[c]{Earth\\Observation}
& \makecell[c]{Spatiotemporal\\Metadata}
& \makecell[c]{Observation\\Environment}
& \makecell[c]{Fine-grained\\Text} \\
\midrule
EarthDial-Dataset~\cite{soni2025earthdial}     & 11.1M & \textcolor{gray}{\scalebox{1.3}{\checkmark}} & \textcolor{gray}{\scalebox{1.3}{\checkmark}} & -        & -         & - \\
TEOChatlas~\cite{irvin2024teochat}             & 554K  & \textcolor{gray}{\scalebox{1.3}{\checkmark}} & \textcolor{gray}{\scalebox{1.3}{\checkmark}} & -        & -         & - \\
FIT-RS~\cite{luo2024skysensegpt}               & 1.8M   & \textcolor{gray}{\ding{55}}   & \textcolor{gray}{\scalebox{1.3}{\checkmark}} & -        & -         & - \\
MMRS-1M~\cite{zhang2024earthgpt}               & 1.0M   & \textcolor{gray}{\ding{55}}  & \textcolor{gray}{\scalebox{1.3}{\checkmark}}   & -        & -         & - \\
Git-10M~\cite{liu2025text2earth}               & 10M   & -         & -         & \textcolor{gray}{\ding{55}} & \textcolor{gray}{\scalebox{1.3}{\checkmark}} & \textcolor{gray}{\ding{55}} \\
Street2Sat-Text~\cite{ye2025satellite}         & 72K  & -         & -         & \textcolor{gray}{\ding{55}} & \textcolor{gray}{\scalebox{1.3}{\checkmark}} & \textcolor{gray}{\scalebox{1.3}{\checkmark}} \\
CVACT-Text~\cite{ye2025satellite}              & 88K  & -         & -         & \textcolor{gray}{\ding{55}} & \textcolor{gray}{\scalebox{1.3}{\checkmark}} & \textcolor{gray}{\scalebox{1.3}{\checkmark}} \\
\rowcolor{gray!15}
\textbf{\data}                              & 1.1M &\textcolor{gray}{\scalebox{1.3}{\checkmark}}  & \textcolor{gray}{\scalebox{1.3}{\checkmark}} & \textcolor{gray}{\scalebox{1.3}{\checkmark}} & \textcolor{gray}{\scalebox{1.3}{\checkmark}} & \textcolor{gray}{\scalebox{1.3}{\checkmark}}
\\
\bottomrule
\end{tabular}%
}
\vspace{-0.3em}
\end{table}


\begin{figure*}[t]
  \centering
  \includegraphics[width=\textwidth]{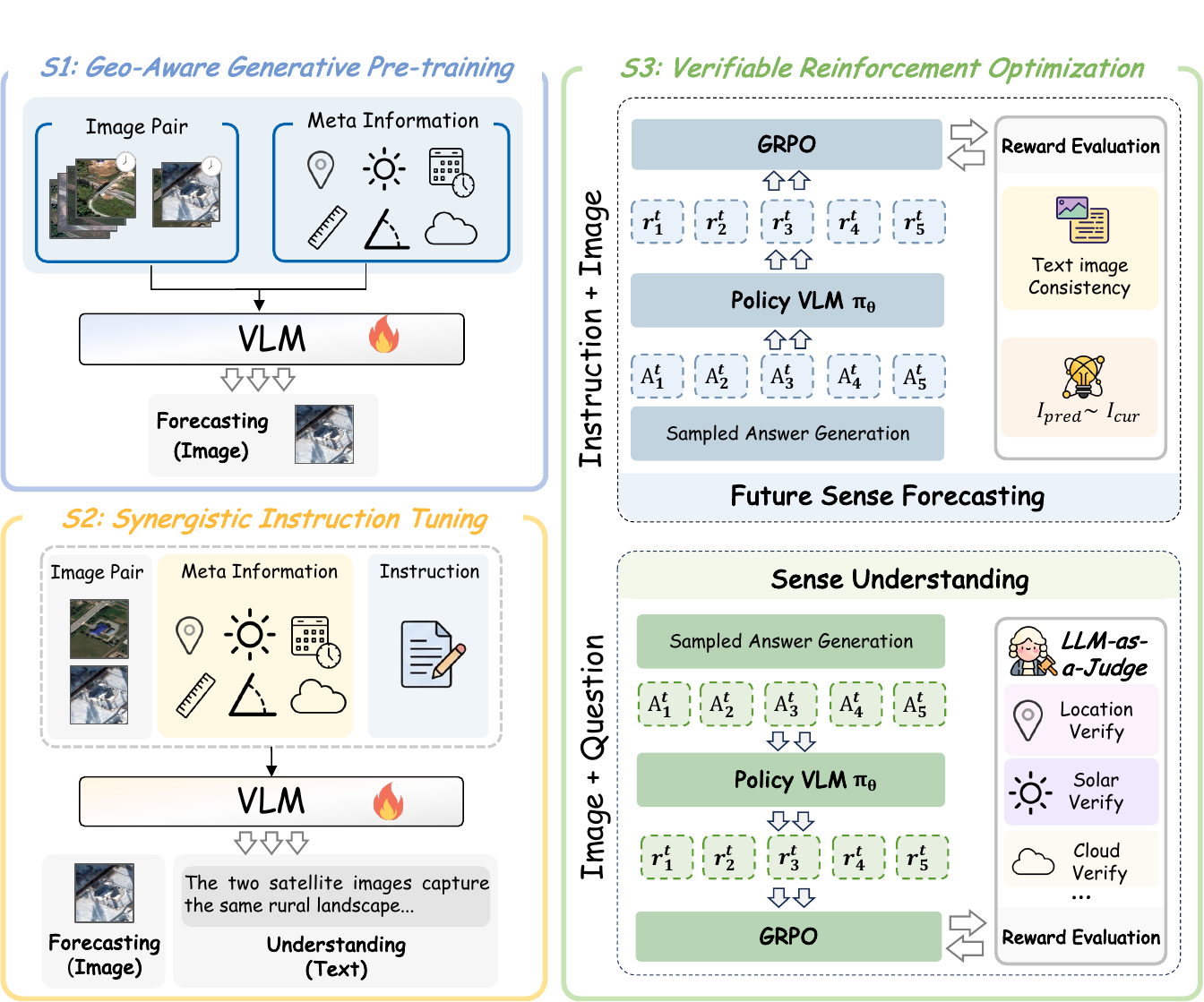}
\caption{\textbf{Overview of \name{}.} The framework is a vision-language world model trained via a three-stage pipeline: \textbf{S1:}~geo-aware generative pre-training on metadata-conditioned image forecasting, \textbf{S2:}~synergistic instruction tuning for joint understanding and forecasting, and \textbf{S3:}~verifiable reinforcement optimization with task-specific rewards.}
  \label{fig:method}
\vspace{-0.3em}
\end{figure*}

\section{Method}
\label{sec:method}

\subsection{Preliminary}
\label{subsec:preliminary}

\noindent\textbf{Problem Definition.}
Let \(I\) denote a remote sensing image and \(m\) its associated geospatial metadata (e.g., coordinates, ground sampling distance, timestamp, sun angles, and cloud statistics). We formulate both \textit{Spatiotemporal Change Question-Answering} (ST-CQA) and \textit{Text-Guided Future Scene Forecasting} (TFSF) as instruction-conditioned sequence generation tasks. Given a prompt \(P\) containing image placeholders \texttt{<image>} and the corresponding metadata \(m\), the objective is to model the conditional probability of the output sequence \(y\):
\begin{equation}
    p_{\theta}(y \mid P, I, m).
\end{equation}
For ST-CQA, \(y\) consists of natural language tokens; for TFSF, \(y\) consists of discrete visual tokens.

\noindent\textbf{Unified Tokenization and Objective.}
We employ a MoVQGAN~\cite{zheng2022movq} tokenizer (codebook size \(K=16{,}384\), sequence length \(L=1{,}024\)) to convert each image \(I\) (256 $\times$ 256) into discrete visual tokens \(z = \mathrm{Tok}(I)\). Both text and visual token generation are treated as a single autoregressive task. The model is trained with next-token prediction on the mixed-modality sequence \(s\):
\begin{equation}
    \mathcal{L}_{\mathrm{AR}}(\theta) = - \sum_{i=1}^{T} \log p_{\theta}(s_i \mid s_{<i}, P, m),
\end{equation}
where \(s_i\) is either a text or visual token. At inference, visual tokens are decoded as \(\hat{I} = \mathrm{Dec}(z)\).

\noindent\textbf{Task-Specific Prompts.}
The model receives textual prompts that combine visual observations, geospatial metadata, and task-specific language.

For Text-Guided Future Scene Forecasting (TFSF), the prompt includes the current observation \((I_{\mathrm{cur}}, m_t)\), a natural-language instruction \(T_{\mathrm{ins}}\) describing the desired changes, and target metadata \(m_{t'}\):
\begin{equation}
    P_{\mathrm{TFSF}} = \{ \mathcal{I}_{\mathrm{cur}},\ T_{\mathrm{ins}},\ m_t,\ m_{t'} \}.
\end{equation}
For geo-aware generative pre-training, we use a simplified text-free version:
\begin{equation}
    P_{\mathrm{FSF}} = \{ \mathcal{I}_{\mathrm{cur}},\ m_t,\ m_{t'} \}.
\end{equation}

For Spatiotemporal Change Question-Answering (ST-CQA), the prompt consists of a natural-language question \(Q\) about spatiotemporal changes, the bi-temporal pair \((I_{\mathrm{pre}}, I_{\mathrm{post}})\), and the corresponding metadata:
\begin{equation}
    P_{\mathrm{ST\text{-}CQA}} = \{ \mathcal{I}_{\mathrm{pre}},\ \mathcal{I}_{\mathrm{post}},\ Q,\ m_{\mathrm{pre}},\ m_{\mathrm{post}} \}.
\end{equation}
Conditioning on both metadata and task-specific text enables the model to separate physical land-cover changes from sensor-induced variations while following user intent.

\subsection{\name{}: A Unified World Model for Remote Sensing}
\label{subsec:worldmodel}

\name{} is a unified world model designed to perceive, understand, and forecast the spatiotemporal dynamics of Earth’s surface from satellite imagery. Unlike conventional vision-language models trained primarily on natural scenes, \name{} explicitly encodes the physical rules that govern remote sensing observations—including sun angles, atmospheric conditions, land-cover evolution, and acquisition-time variations within a single autoregressive framework.

Built upon Qwen3-VL-2B-Instruct with only 2B parameters, \name{} encodes satellite images into visual tokens, fuses them with geospatial metadata, and autoregressively produces mixed-modality outputs: natural-language responses for ST-CQA or discrete visual tokens for future scene forecasting. By treating understanding and forecasting as instances of the same next-token prediction objective in a shared latent space, \name{} establishes a bidirectional connection between perception and simulation. This unified formulation bridges perception and simulation to advance remote sensing intelligence.

\subsection{Learning Remote Sensing World Dynamics}
\label{subsec:learning}

To instill robust physical and semantic priors, \name{} is trained through three complementary objectives: (1) Geo-Aware Generative Pre-training (GAGP) conditions forecasting on geographic and acquisition metadata; (2) synergistic instruction tuning (SIT) jointly trains understanding and forecasting; and (3) verifiable reinforcement optimization (VRO) that refines outputs with verifiable, task-specific rewards. These objectives progressively build world-modeling capabilities from low-level physical simulation to high-level task alignment(\cref{fig:method}).

\noindent\textbf{Geo-Aware Generative Pre-training (GAGP)} performs purely generative pre-training on multi-temporal image sequences without any textual descriptions or language supervision. For each geographic location, we sample a source observation \((I_{\mathrm{cur}}, m_t)\) and a corresponding target observation \((I_{t'}, m_{t'})\). The model is conditioned exclusively on geospatial metadata using the text-free forecasting prompt \(P_{\mathrm{FSF}}\) to autoregressively predict the target visual token sequence \(z_{t'} = \mathrm{Tok}(I_{t'})\):
\begin{equation}
\mathcal{L}_{\mathrm{GAGP}}(\theta)
= -\mathbb{E}\left[\sum_{i=1}^{|z_{t'}|}\log p_{\theta}\!\left(z_{t',i} \mid z_{t',<i}, P_{\mathrm{FSF}}\right)\right].
\end{equation}
This objective enables the model to condition future scene forecasting directly on geographic and acquisition metadata.

\noindent\textbf{Synergistic instruction tuning (SIT)} performs joint instruction tuning on a mixed dataset \(\mathcal{D}^{\mathrm{SIT}} = \mathcal{D}_{\mathrm{ST\text{-}CQA}} \cup \mathcal{D}_{\mathrm{TFSF}}\). Regardless of output modality (text or visual tokens), the unified next-token prediction objective is optimized:
\begin{equation}
\mathcal{L}_{\mathrm{SIT}}(\theta)
= -\mathbb{E}_{(P,y)\sim \mathcal{D}^{\mathrm{SIT}}}\left[\sum_{i=1}^{|y|}\log p_{\theta}\!\left(y_i \mid y_{<i}, P\right)\right].
\end{equation}
Prompts are carefully enriched: TFSF prompts incorporate textual constraints to guide specific land-cover transitions, while ST-CQA prompts demand detailed descriptions of both changed and unchanged elements together with explicit reasoning about sensor-induced variations. This synergistic training creates a closed feedback loop that simultaneously improves forecasting controllability and semantic fidelity in understanding.

\noindent\textbf{Verifiable reinforcement optimization (VRO)} refines the SIT policy using Group Relative Policy Optimization (GRPO)~\cite{guo2025deepseek,shao2024deepseekmath} without requiring a separate value network. The optimization operates on both tasks and employs task-specific rewards derived directly from reference signals and prompt metadata—via cosine similarity for TFSF and an LLM judge for ST-CQA—rather than learned reward models, thereby minimizing reward hacking and ensuring reliable alignment.

For the Text-Guided Future Scene Forecasting (TFSF) task, the model outputs a predicted visual-token sequence \(z_{\mathrm{pred}}\in\{1,\dots,K\}^{L}\). These tokens are decoded into pixel space via the frozen decoder to produce the synthesized future image \(\hat{I}=\mathrm{Dec}(z_{\mathrm{pred}})\). The conditioning prompt supplies the current image \(I_{\mathrm{cur}}\) together with the textual instruction \(T_{\mathrm{ins}}\). We compute the similarities using a frozen vision-language embedding model \(f(\cdot)\)~\cite{li2026qwen3}:
\begin{equation}
s_{\mathrm{it}}=\cos\!\left(f(\hat{I}),\,f(T_{\mathrm{ins}})\right),\qquad
s_{\mathrm{ir}}=\cos\!\left(f(\hat{I}),\,f(I_{\mathrm{cur}})\right).
\end{equation}
The final TFSF reward is defined as
\begin{equation}\label{eq:tfsf_reward}
r_{\mathrm{TFSF}} = s_{\mathrm{it}} + \lambda\, s_{\mathrm{ir}},
\end{equation}
where \(\lambda\) balances description faithfulness against spatial consistency with the current image. This formulation acknowledges the non-unique nature of future forecasting by rewarding any plausible, condition-consistent outcome rather than enforcing pixel-level matching to a single ground-truth future scene.

For the Spatiotemporal Change Question-Answering (ST-CQA) task, we evaluate the generated caption \(\hat{y}\) against the ground-truth reference caption \(y\) using an LLM-based judge (Qwen3-30B-A3B-Instruct-2507)~\cite{yang2025qwen3}. The judge receives the full prompt context together with explicitly parsed spatiotemporal and environmental metadata extracted from the input (coordinates, timestamp, viewing geometry, sun angles, cloud cover statistics, etc.). This metadata grounding enables the judge to detect and penalize contradictions with acquisition conditions (e.g., impossible illumination changes) that traditional n-gram metrics would miss. The LLM outputs a scalar quality score in \([0,100]\), which is clipped and normalized to produce the final reward:
\begin{equation}
r_{\mathrm{ST\text{-}CQA}} = \mathrm{clip}\!\left(\frac{\mathrm{score}(\hat{y},y;x)}{100},\,0,\,1\right).
\end{equation}
Compared with BLEU/ROUGE-style overlap metrics, this LLM judge provides semantically richer evaluation of temporal reasoning, change description completeness, and physical plausibility.

The GRPO objective then directly optimizes the policy \(\pi_{\theta}\) by maximizing the group-relative advantage \(A_{\mathrm{grp}}\) computed over sampled completions while applying KL regularization toward the SIT policy:
\begin{equation}
\max_{\theta}\ \mathbb{E}\!\left[A_{\mathrm{grp}}(x,\hat{y})\right]
-\gamma\,\mathrm{KL}\!\left(\pi_{\theta}(\cdot\mid x)\,\|\,\pi_{\theta_0}(\cdot\mid x)\right).
\end{equation}
Collectively, GAGP, SIT, and VRO equip \name{} with a coherent internal world representation of remote sensing dynamics, enabling robust performance on both perception and forecasting tasks.
\section{Experiments}
\label{sec:exp}

\subsection{Experimental Setups}
\label{sec:exp_setup}
\noindent\textbf{Evaluation Benchmarks.}
We evaluate \name{} on two tasks. \textit{Spatiotemporal Change Question-Answering} (ST-CQA) measures how well a model describes observed bi-temporal changes; we report GPT-Score, BLEU-1, METEOR, ROUGE-L, S-BERT, SimCSE, ST5-SCS, and average response length on a 5K subset (\cref{tab:rswm-understanding}). \textit{Text-Guided Future Scene Forecasting} (TFSF) measures whether a model can synthesize a plausible post-temporal image from a text instruction and geographic context; we report FID, CosSim\cite{li2026qwen3}, and four GPT-based scores (Similarity, Quality, OA, AA) on a 1.6K subset (\cref{tab:rswm-generation}).
\noindent\textbf{Baselines.}
For ST-CQA, we compare with closed-source models (GPT-5.1~\cite{openai2025gpt51}, Gemini-3-Flash~\cite{google2025gemini3flash}), generic open-source VLMs spanning 2B--235B (Qwen-VL series~\cite{bai2025qwen3vltechnicalreport}, LLaVA-OV~\cite{an2025llava}, InternVL3.5~\cite{wang2025internvl3}), and two domain-specific remote sensing models (EarthDial-RGB~\cite{soni2025earthdial}, TEOChat~\cite{irvin2024teochat}). For TFSF, baselines include closed-source generators (Gemini-2.5-Flash Image~\cite{comanici2025gemini}, GPT-Image-1.5, GPT-Image-1-mini) and open-source models across different generation paradigms: diffusion-based CRS-Diff~\cite{tang2024crs}, adapter-based SD3.5-Large-IPA~\cite{sd35-large-ipa} and FLUX.1-Kontext~\cite{labs2025flux}, and the unified model BAGEL~\cite{deng2025emerging}.

\noindent\textbf{Implementation Details.}
\name{} builds on Qwen3-VL-2B-Instruct with the vision encoder and multimodal projector frozen throughout all stages. The \textit{GAGP stage} trains on 371K generation samples, the \textit{SIT stage} fine-tunes on 742K generation and understanding samples, and the \textit{VRO stage} applies GRPO on 16K generation and understanding samples with a KL penalty that combines semantic consistency and perceptual quality rewards. All experiments are conducted on 8 NVIDIA A800 (80\,GB) GPUs using DeepSpeed ZeRO-3 and Flash Attention 2. Full hyperparameters are provided in the supplementary material.

\begin{table*}[t]
\centering
\small
\setlength{\tabcolsep}{4pt}
\caption{\textbf{Spatiotemporal change question-answering results on the 5K subset.} The table compares RS-WorldModel with commercial, open-source, and domain-specific baselines. Baseline references are provided in \cref{sec:exp_setup}.}
\label{tab:rswm-understanding}
\resizebox{\textwidth}{!}{%
\begin{tabular}{llcccccccr}
\toprule
& & & \multicolumn{3}{c}{\textbf{N-Gram}} & \multicolumn{3}{c}{\textbf{Contextual Similarity}} & \\
\cmidrule(lr){4-6} \cmidrule(lr){7-9}
\textbf{Method} & \textbf{Size} & \textbf{GPT-S}$\uparrow$ & \textbf{B-1}$\uparrow$ & \textbf{MTR}$\uparrow$ & \textbf{R-L}$\uparrow$ & \textbf{S-BERT}$\uparrow$ & \textbf{SimCSE}$\uparrow$ & \textbf{ST5}$\uparrow$ & \textbf{Len} \\
\midrule
\multicolumn{10}{l}{\textbf{\textit{Closed-Source Model}}} \\
\addlinespace[0.25em]
\color{gray} GPT-5.1~\cite{openai2025gpt51} & \color{gray} - & \color{gray} 91.17 & \color{gray} 16.82 & \color{gray} 20.87 & \color{gray} 14.59 & \color{gray} 77.19 & \color{gray} 78.28 & \color{gray} 76.70 & \color{gray} 817 \\
\color{gray} Gemini-3-Flash~\cite{google2025gemini3flash} & \color{gray} - & \color{gray} 88.02 & \color{gray} 31.75 & \color{gray} 22.49 & \color{gray} 19.64 & \color{gray} 84.31 & \color{gray} 84.27 & \color{gray} 82.22 & \color{gray} 350 \\
\midrule
\multicolumn{10}{l}{\textbf{\textit{Open-Source Model}}} \\
\addlinespace[0.25em]
\color{gray} Qwen3-VL-32B~\cite{bai2025qwen3vltechnicalreport} & \color{gray} 32B & \color{gray} 87.79 & \color{gray} 33.41 & \color{gray} 25.25 & \color{gray} 21.67 & \color{gray} 87.11 & \color{gray} 84.95 & \color{gray} 84.10 & \color{gray} 385 \\
\color{gray}InternVL3.5-38B~\cite{wang2025internvl3}& \color{gray}38B & \color{gray}83.44 & \color{gray}37.80 & \color{gray}18.94 & \color{gray}19.72 & \color{gray}81.74 & \color{gray}79.97 & \color{gray}79.30 & \color{gray}237 \\
\color{gray} Qwen2.5-VL-72B~\cite{bai2025qwen2} & \color{gray} 72B & \color{gray} 86.40 & \color{gray} 37.06 & \color{gray} 19.78 & \color{gray} 19.83 & \color{gray} 84.30 & \color{gray} 82.11 & \color{gray} 81.68 & \color{gray} 310 \\
\color{gray} Qwen3-VL-235B-A22B~\cite{bai2025qwen3vltechnicalreport} & \color{gray} 235B & \color{gray} 87.64 & \color{gray} 31.25 & \color{gray} 24.35 & \color{gray} 20.22 & \color{gray} 83.10 & \color{gray} 83.48 & \color{gray} 81.90 & \color{gray} 406 \\
\noalign{\vskip 2pt} 
\cdashline{1-10}
\noalign{\vskip 3pt}
\multirow{3}{*}{Qwen3-VL~\cite{bai2025qwen3vltechnicalreport}} 
& 2B & 75.14 & 36.79 & 19.01 & 21.71 & 79.47 & 78.10 & 77.46 & 257 \\
& 4B & 80.85 & 34.26 & 22.44 & 21.75 & 80.76 & 79.70 & 78.01 & 334 \\
& 8B & 76.79 & 39.07 & 19.68 & 20.35 & 80.08 & 78.43 & 78.78 & 238 \\
\noalign{\vskip 2pt} 
\cdashline{1-10}
\noalign{\vskip 3pt}
\multirow{2}{*}{LLaVA-OV-1.5~\cite{an2025llava}} 
& 4B & 65.85 & 36.70 & 15.90 & 18.79 & 75.52 & 76.26 & 75.00 & 183 \\
& 8B & 68.96 & 39.71 & 17.09 & 18.89 & 77.54 & 76.91 & 77.14 & 202 \\
\noalign{\vskip 2pt} 
\cdashline{1-10}
\noalign{\vskip 3pt}
\multirow{4}{*}{InternVL3.5~\cite{wang2025internvl3}} 
& 2B & 72.41 & 31.02 & 16.60 & 17.13 & 77.87 & 75.81 & 75.56 & 259 \\
& 4B & 78.90 & 34.76 & 18.57 & 18.42 & 79.14 & 77.43 & 77.26 & 255 \\
& 8B & 77.05 & 35.26 & 18.21 & 18.02 & 79.19 & 77.61 & 77.30 & 245 \\
& 14B & 80.67 & 34.87 & 19.42 & 19.18 & 80.76 & 79.33 & 78.67 & 263 \\
\noalign{\vskip 2pt} 
\cdashline{1-10}
\noalign{\vskip 3pt} 
EarthDial-RGB~\cite{soni2025earthdial} & 4B & 17.51 & 0.00 & 0.97 & 3.12 & 29.34 & 31.15 & 38.76 & 10 \\
TEOChat~\cite{irvin2024teochat} & 7B & 36.85 & 0.03 & 2.99 & 7.39 & 52.38 & 55.85 & 50.10 & 24 \\
\midrule
\rowcolor{gray!15}
RS-WorldModel & 2B & \textbf{86.20} & \textbf{50.59} & \textbf{22.50} & \textbf{26.35} & \textbf{90.45} & \textbf{86.75} & \textbf{88.32} & 207 \\
\bottomrule
\end{tabular}%
}
\vspace{-0.3em}
\end{table*}
\subsection{Main Results}
\noindent\textbf{Quantitative Results.}
We report results on both tasks below.

\noindent\textbf{(1) \underline{Understanding.}}
\cref{tab:rswm-understanding} reports ST-CQA results. With only 2B parameters, \name{} ranks first among all open-source baselines on BLEU-1, ROUGE-L, and all three contextual similarity metrics. The gain over the same-scale Qwen3-VL-2B is substantial: ROUGE-L improves by 21\% and S-BERT by 14\%. \name{} also surpasses models 16--120$\times$ larger on most metrics, e.g., Qwen3-VL-32B scores 84.10 on ST5-SCS, while \name{} reaches 88.32.
We attribute this to the three-stage training pipeline. Domain-specific pre-training on 371K remote sensing generation samples (GAGP) anchors temporal reasoning in geospatial context, a capability absent from off-the-shelf VLMs regardless of scale. Joint instruction tuning (SIT) then transfers generation-side spatial knowledge to the understanding task, improving caption completeness. The RL stage (VRO) further refines outputs via a judge-based reward that penalizes metadata-inconsistent descriptions.

Two domain-specific baselines, EarthDial-RGB and TEOChat, score below 40 on GPT-Score, indicating that existing remote sensing models are not designed for open-ended temporal captioning. Among closed-source models, GPT-5.1~\cite{openai2025gpt51} achieves the highest GPT-Score but produces responses averaging 817 tokens (nearly 4$\times$ the length of \name{}) with lower n-gram and contextual similarity scores, suggesting verbose but less precise descriptions.

\begin{table}[t]
\centering
\footnotesize
\setlength{\tabcolsep}{3pt}
\caption{\textbf{Text-guided future scene forecasting results on the 1.6K subset.} Baseline references are provided in \cref{sec:exp_setup}.}
\label{tab:rswm-generation}
\begin{tabular}{llcccccc}
\toprule
\multirow{2}{*}{\textbf{Method}} & \multirow{2}{*}{\textbf{Size}} & \multirow{2}{*}{\textbf{FID}$\downarrow$} & \multirow{2}{*}{\textbf{CosSim}$\uparrow$} & \multicolumn{4}{c}{\textbf{GPT Scores}$\uparrow$} \\
\cmidrule(l){5-8}
 & & & & Sim. & Qual. & OA & AA \\
\midrule
\multicolumn{8}{l}{\textbf{\textit{Closed-Source Model}}} \\
\addlinespace[0.25em]
\color{gray}Gemini-2.5-Flash Image~\cite{comanici2025gemini} & \color{gray}- & \color{gray}46.14 & \color{gray}69.21 & \color{gray}46.63 & \color{gray}46.95 & \color{gray}93.58 & \color{gray}46.79 \\
\color{gray}GPT-Image-1.5~\cite{openai2025gptimage15_model} & \color{gray}- & \color{gray}83.51 & \color{gray}66.05 & \color{gray}46.94 & \color{gray}47.06 & \color{gray}94.00 & \color{gray}47.00 \\
\color{gray}GPT-Image-1-mini~\cite{openai2025gptimage1mini} & \color{gray}- & \color{gray}92.27 & \color{gray}65.95 & \color{gray}44.76 & \color{gray}45.96 & \color{gray}90.72 & \color{gray}45.36 \\
\midrule
\multicolumn{8}{l}{\textbf{\textit{Open-Source Model}}} \\
\addlinespace[0.25em]
CRS-Diff ~\cite{tang2024crs} & 0.9B & 82.76 & 63.09 & 27.04 & 30.97 & 58.01 & 29.01 \\
BAGEL ~\cite{deng2025emerging} & 7B & 78.47 & 62.82 & 44.25 & 42.13 & 86.38 & 43.19 \\
SD3.5-Large-IPA ~\cite{sd35-large-ipa} & 8B & 97.88 & 66.69 & 33.15 & 40.63 & 73.78 & 36.89 \\
FLUX.1-Kontext ~\cite{labs2025flux} & 12B & 81.92 & 64.67 & 39.00 & 42.41 & 81.41 & 40.70 \\
\rowcolor{gray!15}
RS-WorldModel & 2B & \textbf{43.13} & \textbf{68.34} & \textbf{44.59} & \textbf{44.84} & \textbf{89.43} & \textbf{44.71} \\
\bottomrule
\end{tabular}
\vspace{-0.3em}
\end{table}

\noindent\textbf{(2) \underline{Forecasting.}}
\cref{tab:rswm-generation} reports TFSF results. \name{} ranks first among all open-source models on every metric, reducing FID by 48\% relative to CRS-Diff and by 47\% relative to FLUX.1-Kontext while attaining the highest CosSim and GPT scores.
A comparison across generation paradigms reveals distinct trade-offs. CRS-Diff, a diffusion model conditioned on change instructions, produces perceptually reasonable images but scores lowest on Similarity, suggesting limited adherence to the textual change description. BAGEL, a unified model like ours, scores competitively on Similarity (44.25) but incurs a substantially higher FID (78.47), indicating text-faithful yet perceptually weaker outputs. \name{} balances both objectives: its autoregressive formulation with VRO-based reward optimization jointly encourages text faithfulness via $s_{\mathrm{it}}$ and perceptual realism via $s_{\mathrm{ir}}$.
\name{} even surpasses the closed-source Gemini-2.5-Flash Image on FID (43.13 vs.\ 46.14). GPT-Image-1.5 leads on Similarity and OA but with an FID nearly double that of \name{}, reflecting higher text adherence at the cost of perceptual fidelity.


\begin{table*}[t]
\centering
\small
\setlength{\tabcolsep}{4pt}
\caption{\textbf{Ablation on the reference-adherence weight $\lambda$.} Forecasting on the 1.6K subset; understanding on the 5K subset.}
\label{tab:ablation-lambda}
\resizebox{\textwidth}{!}{%
\begin{tabular}{l cc cccc ccc cccc c}
\toprule
& \multicolumn{6}{c}{\textbf{Forecasting (TFSF)}} & \multicolumn{7}{c}{\textbf{Understanding (ST-CQA)}} \\
\cmidrule(lr){2-7} \cmidrule(lr){8-14}
& & & \multicolumn{4}{c}{\textbf{GPT Scores$\uparrow$}} & & \multicolumn{3}{c}{\textbf{N-Gram}} & \multicolumn{3}{c}{\textbf{Contextual Sim.}} & \\
\cmidrule(lr){4-7} \cmidrule(lr){9-11} \cmidrule(lr){12-14}
\textbf{$\lambda$} & \textbf{FID}$\downarrow$ & \textbf{CosSim}$\uparrow$ & Sim. & Qual. & OA & AA & \textbf{GPT-S}$\uparrow$ & \textbf{B-1}$\uparrow$ & \textbf{MTR}$\uparrow$ & \textbf{R-L}$\uparrow$ & \textbf{S-BERT}$\uparrow$ & \textbf{SimCSE}$\uparrow$ & \textbf{ST5}$\uparrow$ & \textbf{Len} \\
\midrule

0.0 & 44.19 & 67.22 & 44.40 & 44.34 & 88.75 & 44.37 & 86.03 & 49.78 & 22.62 & 26.05 & 90.44 & 86.32 & 88.07 & 211 \\
0.1 & 43.64 & 67.24 & 44.23 & 44.41 & 88.63 & 44.32 & 86.05 & 49.61 & \textbf{22.79} & 26.04 & 90.44 & 86.47 & 88.07 & 214 \\
\rowcolor{gray!15}
0.2 & \textbf{43.13} & \textbf{68.34} & \textbf{44.59} & \textbf{44.84} & \textbf{89.43} & \textbf{44.71} & \textbf{86.20} & \textbf{50.59} & 22.50 & \textbf{26.35} & \textbf{90.45} & \textbf{86.75} & \textbf{88.32} & 207 \\
\bottomrule
\end{tabular}%
}
\vspace{-0.3em}
\end{table*}

\noindent\textbf{Qualitative Results.}
To qualitatively evaluate RS-WorldModel’s capabilities in both understanding and forecasting, we present representative examples from the two core tasks.
\FloatBarrier
\begin{wrapfigure}{r}{0.5\textwidth}
    \vspace{-0.8em}
    \centering
    \includegraphics[width=\linewidth]{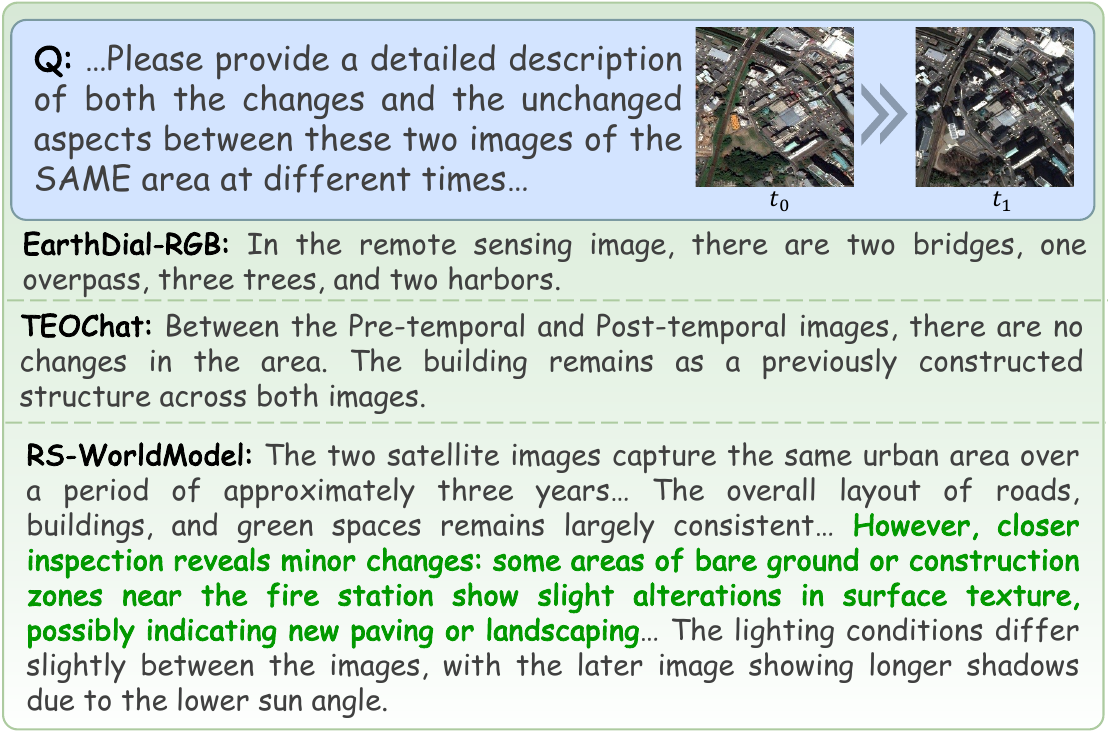}
    \caption{Qualitative comparison on temporal change understanding.}
    \label{fig:qual_understanding}
    \vspace{-1.0em}
\end{wrapfigure}
\noindent\textbf{(1) \underline{Understanding.}}
In the change-understanding scenario (Figure~\ref{fig:qual_understanding}), given a pair of high-resolution satellite images of the same urban area captured approximately three years apart, RS-WorldModel accurately reports the overall layout consistency while identifying subtle surface-texture changes near the fire station and correctly attributing differences in shadow length and orientation to variations in sun elevation and acquisition time. In contrast, several strong baselines either overlook all changes or hallucinate major structural modifications.

\begin{figure}[!htb]
    \centering
    \includegraphics[width=1.0\textwidth]{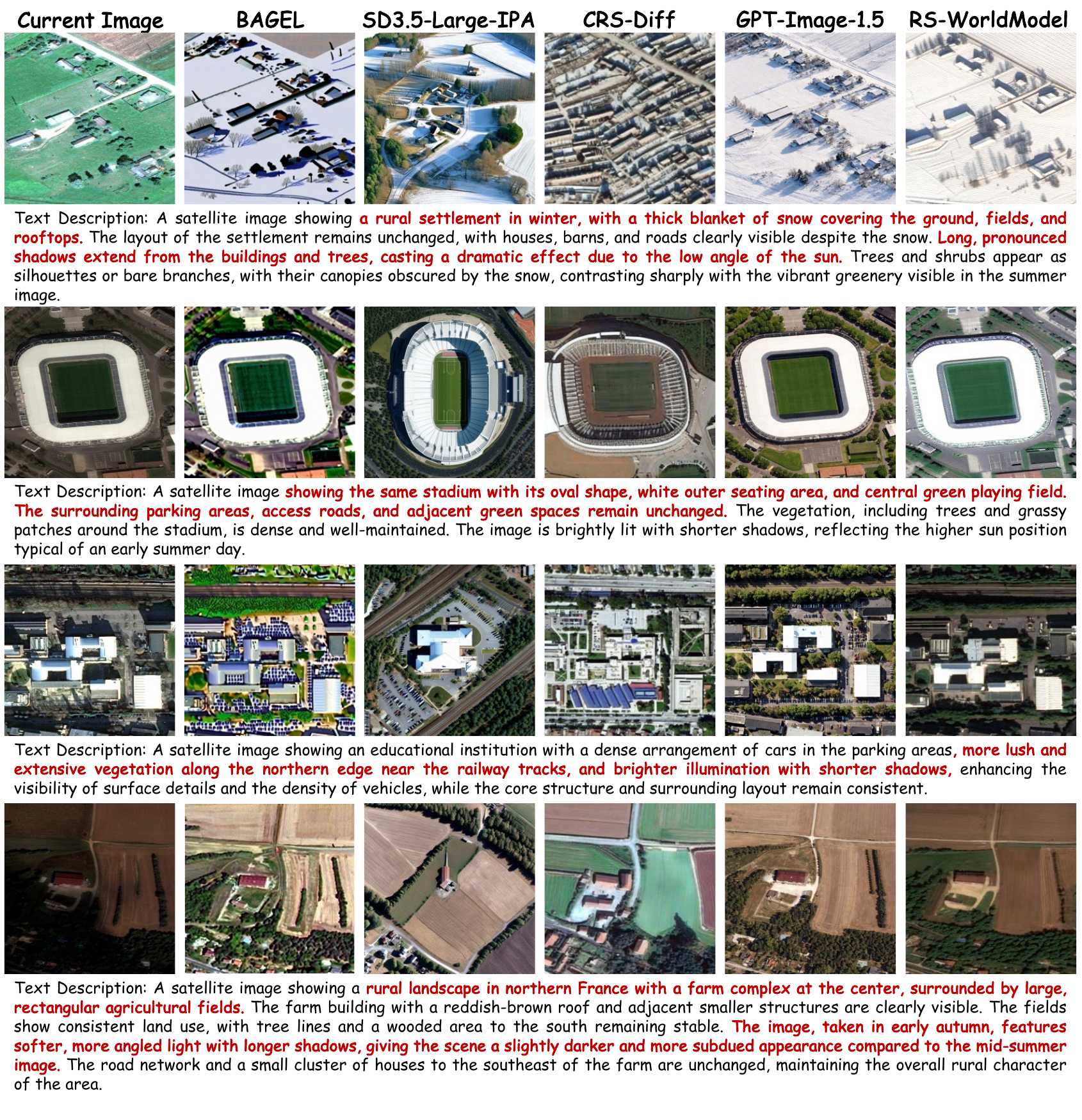}
    \caption{Qualitative comparison on the text-guided satellite image forecasting task. Given detailed textual prompts, RS-WorldModel generates images with superior structural fidelity, shadow consistency, and scene realism compared to strong baselines.}
    \label{fig:qual_generation}
\vspace{-0.3em}
\end{figure}

\noindent\textbf{(2) \underline{Forecasting.}}
In the text-guided forecasting scenario (Figure~\ref{fig:qual_generation}), when conditioned on detailed textual descriptions of recreational and commercial scenes, RS-WorldModel produces photorealistic satellite imagery that faithfully preserves tennis-court layouts, parking configurations, vegetation density, building rooftops, shadow directions, and atmospheric lighting outperforming competing diffusion and autoregressive models in structural fidelity and physical consistency.

\subsection{Ablation Study}
\label{sec:ablation}
\noindent\textbf{Effect of $\lambda$ in the TFSF Reward.}
The hyperparameter $\lambda$ in \cref{eq:tfsf_reward} balances reference-image consistency ($s_{\mathrm{ir}}$) against text-description faithfulness ($s_{\mathrm{it}}$) in the VRO reward. We sweep $\lambda \in \{0.0, 0.1, 0.2\}$ and report results on both tasks (\cref{tab:ablation-lambda}). When $\lambda{=}0$, the reward ignores the reference image entirely, relying on the textual description alone.
On TFSF, increasing $\lambda$ consistently improves all metrics: CosSim rises from 67.22 to 68.34, FID drops from 44.19 to 43.13, and GPT-based OA climbs from 88.75 to 89.43. This confirms that the reference image supplies valuable spatial priors, including building layouts, road networks, and land-cover distributions that anchor structural plausibility beyond what text alone can convey.
On ST-CQA, a consistent trend emerges: GPT-Score improves from 86.03 to 86.20 and BLEU-1 from 49.78 to 50.59 as $\lambda$ increases, with contextual similarity metrics following the same upward pattern. Only METEOR marginally favors $\lambda{=}0.1$ (22.79 vs.\ 22.50). Overall, moderate reference adherence ($\lambda{=}0.2$) uniformly outperforms both the text-only baseline ($\lambda{=}0$) and the weaker reference signal ($\lambda{=}0.1$). We therefore adopt $\lambda{=}0.2$ for all experiments.

\noindent\textbf{Ablation on Training Stages.}
We ablate each training stage on the TFSF task (\cref{tab:training_phase_ablation}). Training with SIT alone (no generative pre-training) yields an FID of 73.55, a Similarity score of 39.78, and an OA of 77.43. Adding GAGP before SIT drops FID to 44.23 and raises Similarity to 42.38 and OA to 83.20, showing that generative pre-training on geo-conditioned data provides strong spatial priors for the downstream forecasting task.
The VRO stage brings a further improvement: the full three-stage pipeline (GAGP$\to$SIT$\to$VRO) achieves an FID of 43.13, Similarity of 44.59, OA of 89.43, and GPT-S of 86.20, outperforming all partial configurations. GAGP alone already reaches an FID of 50.28, but without SIT the model cannot follow change instructions (GPT scores unavailable). Each stage thus contributes a distinct capability, and removing any one leads to measurable degradation.
\begin{table}[t]
\centering
\footnotesize
\setlength{\tabcolsep}{3pt}
\caption{\textbf{Ablation on the three-stage training paradigm.} Forecasting on the 1.6K subset; understanding on the 5K subset. $^*$GAGP-only uses metadata conditioning without text instructions.}
\label{tab:training_phase_ablation}
\begin{tabular}{ccc ccccc cc}
\toprule
& & & \multicolumn{5}{c}{\textbf{Forecasting (TFSF)}} & \multicolumn{2}{c}{\textbf{ST-CQA}} \\
\cmidrule(lr){4-8} \cmidrule(lr){9-10}
GAGP & SIT & VRO & FID$\downarrow$ & Sim.$\uparrow$ & Qual.$\uparrow$ & OA$\uparrow$ & AA$\uparrow$ & GPT-S$\uparrow$ & Len \\
\midrule
$\times$ & $\checkmark$ & $\times$ & 73.55 & 39.78 & 37.64 & 77.43 & 38.71 & 85.63 & 214 \\
$\checkmark$ & $\times$ & $\times$ & 50.28\rlap{$^*$} & -- & -- & -- & -- & -- & -- \\
$\checkmark$ & $\checkmark$ & $\times$ & 44.23 & 42.38 & 40.82 & 83.20 & 41.56 & 85.24 & 201 \\
\rowcolor{gray!15}
$\checkmark$ & $\checkmark$ & $\checkmark$ & \textbf{43.13} & \textbf{44.59} & \textbf{44.84} & \textbf{89.43} & \textbf{44.71} & \textbf{86.20} & 208 \\
\bottomrule
\end{tabular}
\vspace{-0.3em}
\end{table}

\needspace{6\baselineskip}
\begin{wraptable}{r}{0.38\textwidth}
\vspace{-2em}
\centering
\footnotesize
\setlength{\tabcolsep}{4pt}
\caption{\textbf{Geo-metadata ablation in GAGP.} FID on the 1.6K subset.}
\label{tab:geo_ablation}
\begin{tabular}{@{}lc@{}}
\toprule
\textbf{Pre-training} & \textbf{FID}$\downarrow$ \\
\midrule
w/o Geo Metadata & 53.72 \\
\rowcolor{gray!15}
w/ Geo Metadata  & \textbf{50.28} \\
\bottomrule
\end{tabular}
\vspace{-1.0em}
\end{wraptable}

\noindent\textbf{Ablation on Geographic Metadata in GAGP.}
We compare two GAGP variants (\cref{tab:geo_ablation}). Without geographic and acquisition metadata conditioning, FID increases from 50.28 to 53.72, confirming that location and sensor information helps the model learn spatially grounded representations during generative pre-training. Qualitatively, we observe that the geo-conditioned model produces land-cover distributions better aligned with the target region, whereas the variant without metadata tends to generate geographically implausible textures. These results suggest that geographic and acquisition metadata serve as an effective spatial prior for the generative pre-training stage.

\section{Conclusion}

We presented \name{}, a unified world model that jointly addresses spatiotemporal change understanding and text-guided future scene forecasting for remote sensing. Together with \data{}, a 1.1M-sample dataset covering both tasks with fine-grained geographic metadata, \name{} is trained via a three-stage pipeline: Geo-Aware Generative Pre-training, synergistic instruction tuning, and verifiable reinforcement optimization. With only 2B parameters, \name{} surpasses open-source models up to 120$\times$ larger on most ST-CQA metrics and outperforms all open-source baselines and the closed-source Gemini-2.5-Flash Image on forecasting FID. Ablations confirm that each training stage contributes a distinct capability and that the verifiable reward design transfers benefits across both tasks.



%
%
\bibliographystyle{splncs04}
\bibliography{main}
\newpage
\renewcommand\thesection{\Alph{section}}
\setcounter{section}{0}


The appendix includes the following sections:
{\hypersetup{linkcolor=black}
\begin{itemize}
    \item \hyperlink{supp:related_work}{Appendix A: Related Work}
    \item \hyperlink{supp:dataset}{Appendix B: Details about RSWBench-1.1M Dataset}
    \item \hyperlink{supp:additional_implementation_details}{Appendix C: Additional Implementation Details}
    \item \hyperlink{supp:case_studies}{Appendix D: Case Studies}
    \item \hyperlink{supp:prompts}{Appendix E: Prompts}
\end{itemize}
}

\hypertarget{supp:related_work}{\section{Related Work}}

\subsection{Unified multimodal understanding and generation}
Recent studies highlight the advantages of unified multimodal models that jointly handle visual understanding and controllable generation within a single autoregressive framework~\cite{wu2025janus,chen2025janus,chen2025blip3}, enabling bidirectional knowledge transfer via shared representations and yielding stronger semantic consistency, generation controllability, and emergent world-modeling capabilities. Methods like Show-o2~\cite{xie2025show} represent visual information through discrete tokenization and train large language models to perform autoregressive next-token prediction. However, they often suffer from insufficient semantic preservation and degraded downstream understanding performance. Alternatives using continuous encoders typically rely on external diffusion models or mismatched objectives~\cite{cui2025emu3}, resulting in complex designs and prohibitive billion-scale pretraining costs. Inspired by FutureSightDrive~\cite{zeng2025futuresightdrive}, we unify spatiotemporal change understanding and text-guided future scene forecasting through a shared tokenizer and a single next-token prediction objective on mixed text-visual sequences, achieving competitive results with only approximately 1\% of the training costs of prior methods~\cite{wu2025janus,sun2025gdiffretro,lu2025dammfnd,dai2025secure}.

\subsection{Vision-language models for Remote Sensing}
Vision-language models for remote sensing have produced several strong understanding oriented approaches~\cite{weng2025vision}. GeoChat~\cite{kuckreja2024geochat} introduces grounded spatial reasoning, while RSGPT~\cite{hu2025rsgpt} establishes a comprehensive benchmark for VQA, captioning, and other understanding tasks. SkyEyeGPT~\cite{zhan2025skyeyegpt} unifies diverse RS tasks via large-scale instruction tuning, and Skysense-o~\cite{zhu2025skysense} pushes toward open-world interpretation with a vision-centric design. EarthGPT~\cite{zhang2024earthgpt} further extends to multisensor comprehension. More recent efforts, such as SAMChat~\cite{koksal2025samchat}, incorporate chain-of-thought reasoning to improve efficiency on small-scale remote sensing images. Nevertheless, these methods focus exclusively on perception and lack native support for controllable future scene generation or unified world modeling.

\subsection{Large-Scale Remote Sensing Vision-Language Datasets}
Large-scale remote sensing vision-language datasets have been developed to support multimodal understanding tasks. VRSBench~\cite{li2024vrsbench} serves as a versatile benchmark for image understanding, SkySenseGPT~\cite{luo2024skysensegpt} provides a fine-grained instruction tuning dataset, RS-GPT4V~\cite{xu2024rs} offers a unified multimodal instruction-following corpus, and SkyScript~\cite{wang2024skyscript} contributes a large and semantically diverse collection. Some works also explore text-guided satellite image synthesis from street-view inputs with fine-grained spatial textual guidance~\cite{ye2025satellite}. However, these datasets primarily focus on understanding tasks such as VQA and captioning, are mostly single-temporal, and provide no native support for controllable generation. In contrast, our RSWBench-1.1M jointly enables spatiotemporal change understanding and text-guided future scene forecasting with rich fine-grained language annotations and detailed geographic metadata.
\hypertarget{supp:dataset}{\section{Details about RSWBench-1.1M Dataset}}

\begin{figure*}[t!]
  \centering
  
  \includegraphics[width=\textwidth]{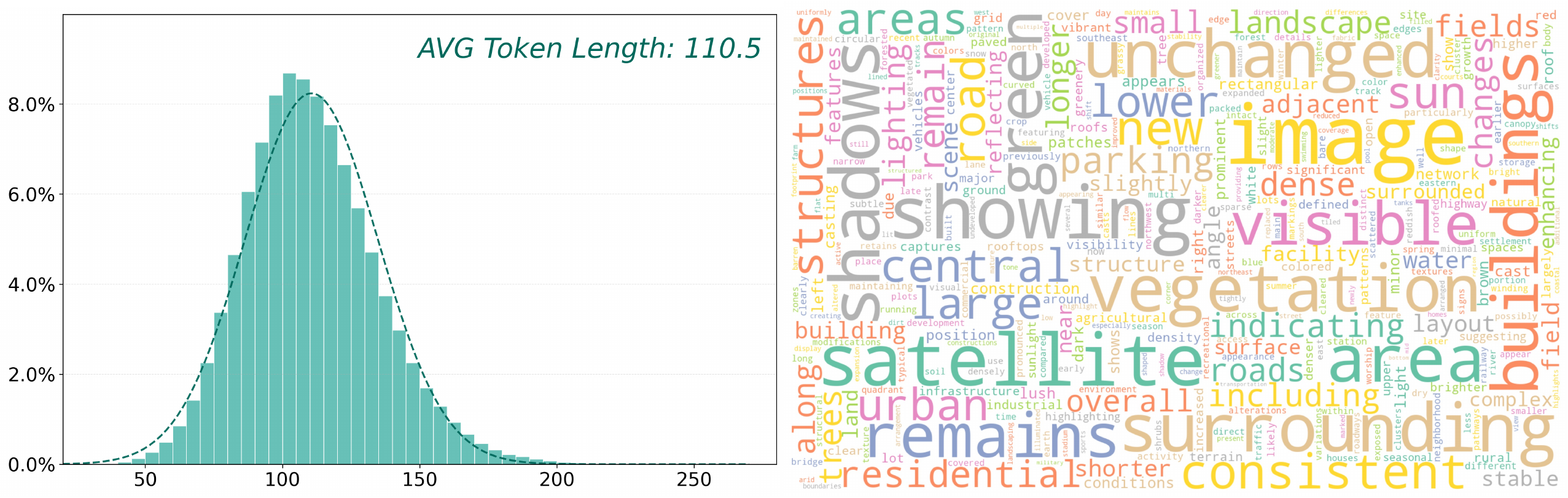}
  \caption{\textbf{Forecasting tasks (TFSF).} Token-length distribution (left) and word cloud of frequent terms (right) for the text-guided future scene forecasting subset.}
  \label{fig:gen_token_wordcloud}
  
  \vspace{1.2em}   
  
  \includegraphics[width=\textwidth]{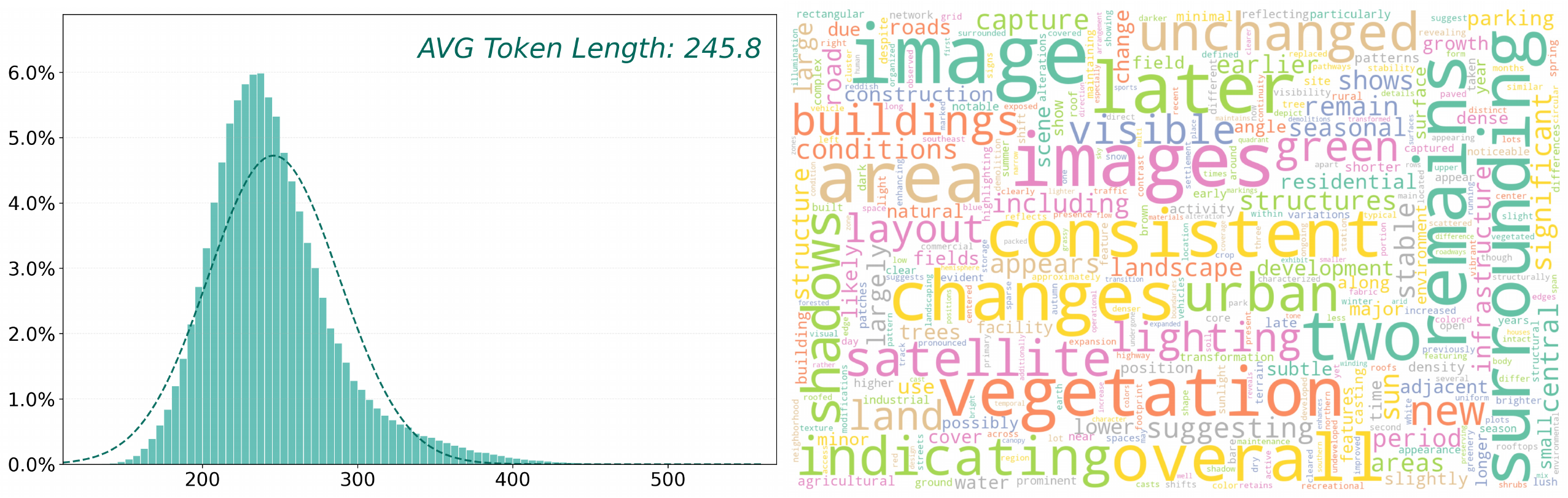}
  \caption{\textbf{Understanding tasks (ST-CQA).} Token-length distribution (left) and word cloud of frequent terms (right) for the spatiotemporal change question-answering subset.}
  \label{fig:under_token_wordcloud}
  
\end{figure*}

To further illustrate the scale and linguistic characteristics of RSWBench-1.1M, we present token-length distributions and word-cloud visualizations for both the forecasting and understanding subsets, as shown in \cref{fig:gen_token_wordcloud,fig:under_token_wordcloud}. These statistics demonstrate the diversity of instructions, the balanced complexity across tasks, and the rich semantic coverage achieved through our automated annotation pipeline.

\hypertarget{supp:additional_implementation_details}{\section{Additional Implementation Details}}

\noindent\textbf{Training details.}
RS-WorldModel is built on Qwen3-VL-2B-Instruct. Across all three stages, we freeze the vision encoder and the multimodal projector and train the remaining parameters in bf16 on 8 NVIDIA A800 GPUs (80\,GB each), using DeepSpeed ZeRO-3 and Flash Attention 2. For Stages~1 and~2, we cap the image resolution at 524,288 pixels and the video resolution at 16,384 pixels, with a context length of 32,768 tokens and a maximum generation length of 2,048 tokens. We further introduce dedicated tokens for geographic coordinates, ground sampling distance, timestamps, sun angles, off-nadir angle, and cloud cover, allowing acquisition metadata to be serialized together with the visual context.

In Stage~1, we carry out geo-aware generative pre-training on 371K forecasting samples for 32 epochs with a per-device batch size of 16 and gradient accumulation of 2. We use a cosine schedule with a peak learning rate of $5\times10^{-4}$ and a warmup ratio of 0.10. Stage~2 starts from the Stage-1 checkpoint and is trained for another 32 epochs on 742K mixed understanding and forecasting samples. The batch configuration remains unchanged, while the peak learning rate is reduced to $1\times10^{-4}$ and the warmup ratio  to 0.02. In both stages, 10\% of the training data is held out for validation; the evaluation batch size is 16, and evaluation is performed every 1000 steps.

In Stage~3, we continue from the Stage-2 checkpoint and apply GRPO on 16K mixed ST-CQA and TFSF samples. For TFSF, we use the reward in Eq.~(\ref{eq:tfsf_reward}) with $\lambda=0.2$, selected by the ablation study in the main paper. For ST-CQA, we adopt Qwen3-30B-A3B-Instruct-2507 as the judge model. We retain KL regularization throughout reinforcement optimization to keep the policy close to the Stage-2 initialization and preserve the instruction-following behavior learned during instruction tuning.

\hypertarget{supp:case_studies}{\section{Case Studies}}

\begin{figure*}[t!]
  \centering
  \includegraphics[width=\textwidth]{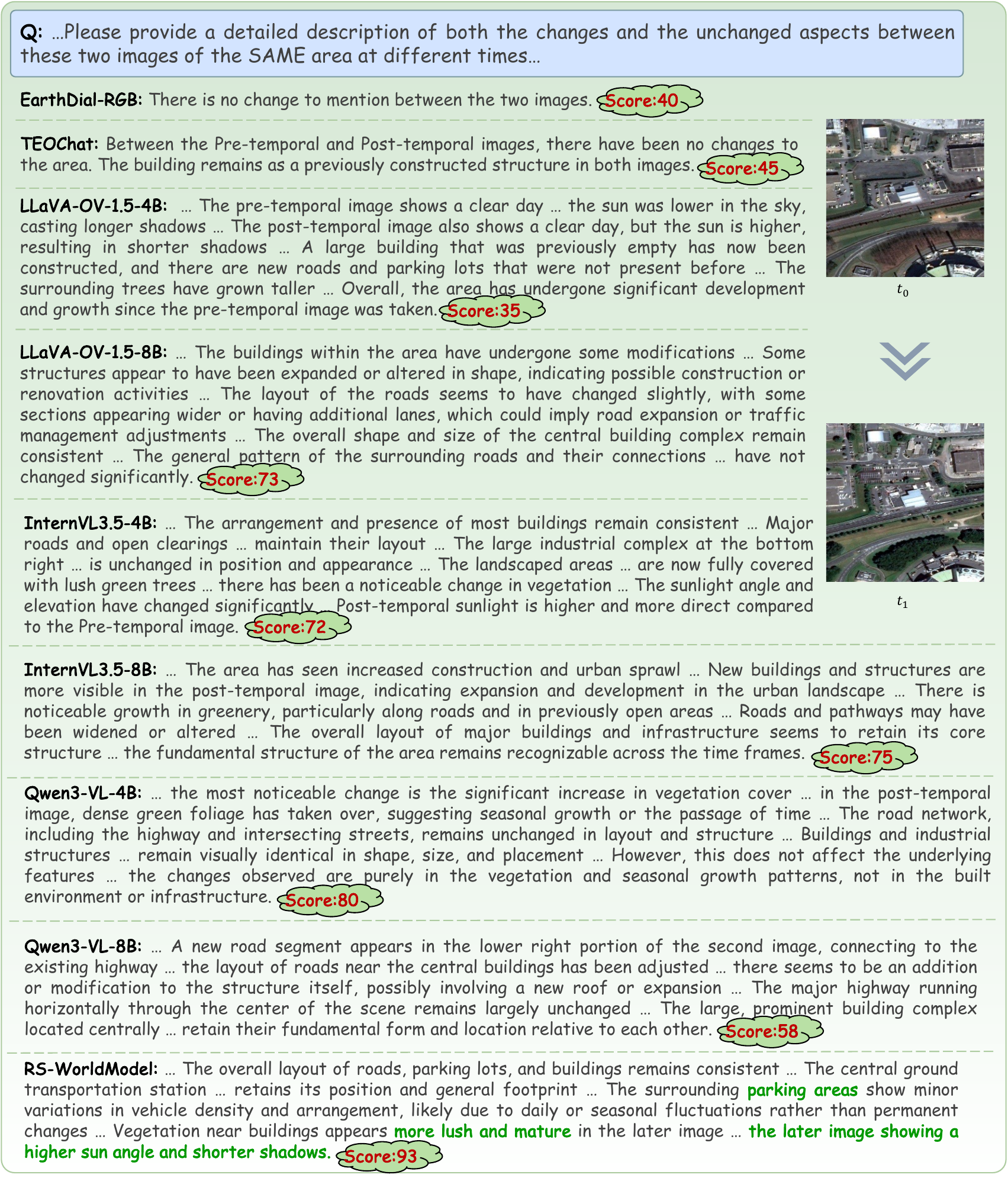}
  \caption{\textbf{ST-CQA case study.} Model responses with GPT-Scores. RS-WorldModel achieves the best score.}
  \label{fig:case-understanding}
\end{figure*}

We present two representative case studies to qualitatively demonstrate RS-WorldModel's superiority in both core tasks. In the spatiotemporal change understanding task (Figure~\ref{fig:case-understanding}), given a bi-temporal pair with the instruction ``Please provide a detailed description of both the changes and the unchanged aspects between these two images of the SAME area at different times'', most baselines either overlook subtle changes or hallucinate major modifications. In contrast, RS-WorldModel accurately identifies layout consistency, vegetation growth, and acquisition-time shadow variations, achieving the highest GPT-Score.

\begin{figure*}[t!]
  \centering
  \includegraphics[width=\textwidth]{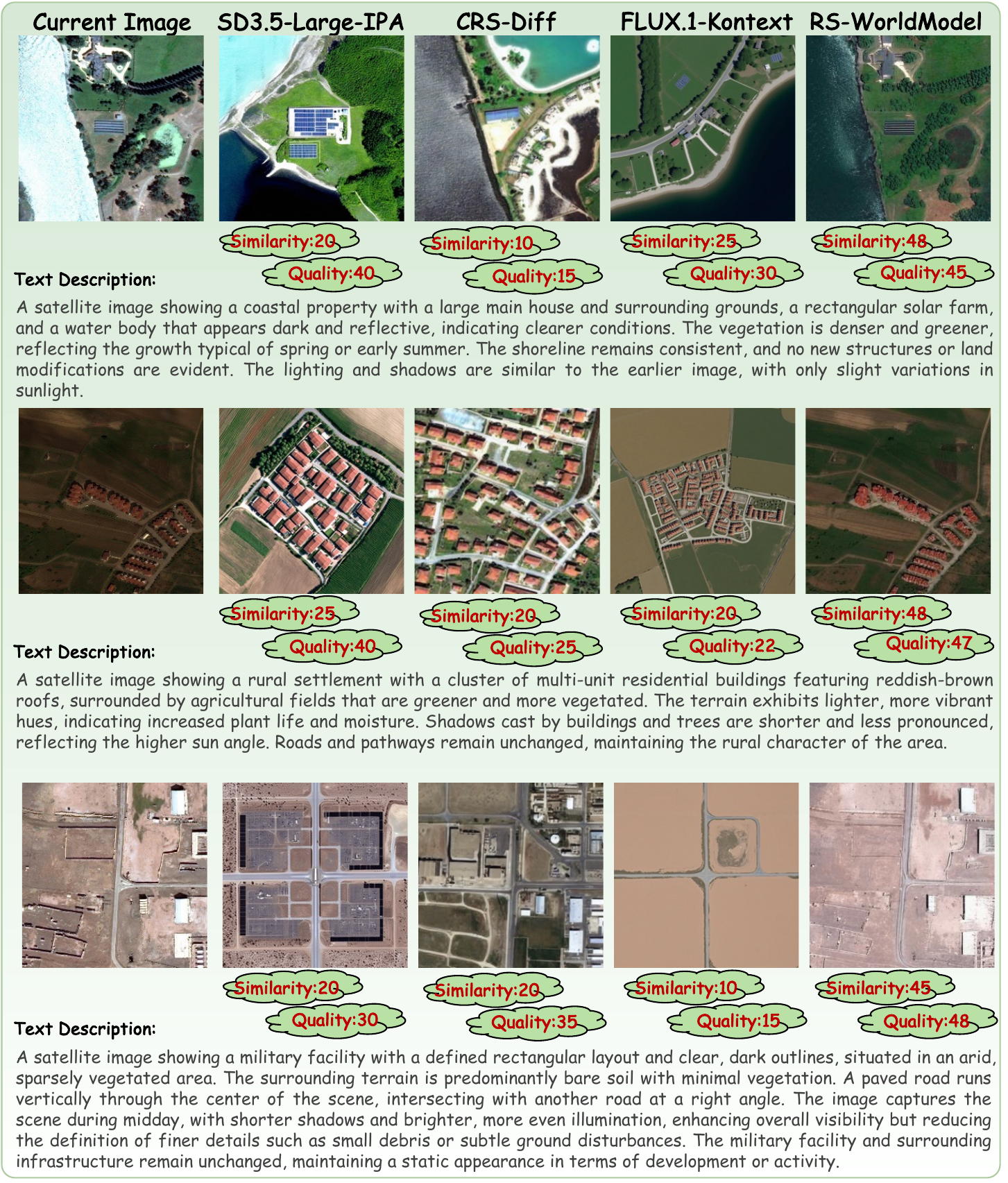}
  \caption{\textbf{TFSF case study.} Generated results for three textual instructions. RS-WorldModel obtains the highest GPT-based scores.}
  \label{fig:case-forecasting}
\end{figure*}

For the text-guided future scene forecasting task (Figure~\ref{fig:case-forecasting}), across three diverse scenarios with identical textual instructions and geographic metadata, RS-WorldModel generates images with superior structural fidelity, shadow consistency, and text adherence, consistently attaining the highest GPT-based Similarity and Quality scores among strong open-source baselines.

\hypertarget{supp:prompts}{\section{Prompts}}

To ensure reproducibility, we present all prompt templates used in our data construction, evaluation, and training pipeline. These prompts are carefully engineered for their respective roles: the \textit{Qwen3-VL-32B Draft Generation Prompt} (~\cref{fig:draft-generation-prompt}) and \textit{Qwen2.5-72B Text Refinement Prompt} (~\cref{fig:text-refinement-prompt}) enable scalable, high-quality annotation of RSWBench-1.1M; the \textit{GPT-5-Nano ST-CQA Scoring Prompt} (~\cref{fig:gpt-5-nano-stcqa-prompt}) and \textit{GPT-4o TFSF Scoring Prompt} (~\cref{fig:gpt-4o-tfsf-prompt}) provide reliable automatic scoring for understanding and generation tasks; the \textit{Qwen3 LLM-as-a-Judge Prompt} (~\cref{fig:qwen3-llm-judge-prompt}) drives verifiable reinforcement optimization (VRO); and the \textit{Stage-1 System Prompt} (~\cref{fig:rs-worldmodel-stage1-prompt}) together with the \textit{Stage-2/3 System Prompt} (~\cref{fig:rs-worldmodel-stage2-3-prompt}) define RS-WorldModel's behavior across training stages. In particular, we adopt a pure LLM as the judge in VRO, inspired by Perception-R1~\cite{xiao2025perception}, rather than a VLM. This design delivers more stable, semantically rich, and metadata-grounded reward signals for geographic and physical plausibility.

\begin{tcolorbox}[
  colback=Black!5,
  colframe=Black!70!Black!70,
  title=Draft Generation Prompt,
  breakable,
  enhanced
]
\small

\begin{tcolorbox}[
  colback=Black!5,
  colframe=Black!50!Black!40,
  boxrule=0.3pt,
  left=2pt,right=2pt,top=1pt,bottom=1pt,
  title=System Prompt
]
You are an advanced AI model capable of combining geospatial metadata to generate temporal captions describing both changes and unchanged aspects between two temporal image phases, as well as constructing text prompts for generating post-temporal images.
\end{tcolorbox}

You will be provided with two satellite images of the same geographic location, centered at coordinates (latitude, longitude) (\texttt{\{center\_lat\_lon\}}), but captured at different times. Geospatial metadata includes:

\begin{itemize}[leftmargin=1.5em, itemsep=0.3em]
  \item \textbf{Acquisition times:} Pre-temporal image at \texttt{\{time[0]\}}, Post-temporal image at \texttt{\{time[1]\}}.
  
  \item \textbf{Bounding boxes for key objects} (normalized coordinates in the range [0, 1000)): Pre-temporal image \texttt{\{"bbox\_2d": \{new\_xyxy\_box[0]\}, "label": "\{category[0]\}"\}}, Post-temporal image \texttt{\{"bbox\_2d": \{new\_xyxy\_box[1]\}, "label": "\{category[1]\}"\}}.
  
  \item \textbf{Sun azimuth:} Pre-temporal image \texttt{\{sun\_azimuth[0]\}} degrees, Post-temporal image \texttt{\{sun\_azimuth[1]\}} degrees.
  
  \item \textbf{Sun elevation:} Pre-temporal image \texttt{\{sun\_elevation[0]\}} degrees, Post-temporal image \texttt{\{sun\_elevation[1]\}} degrees.
  
  \item \textbf{Off-nadir angle:} Pre-temporal image \texttt{\{off\_nadir\_angle[0]\}} degrees, Post-temporal image \texttt{\{off\_nadir\_angle[1]\}} degrees.
  
  \item \textbf{Cloud cover:} Pre-temporal image \texttt{\{cloud\_cover[0]\}}\%, Post-temporal image \texttt{\{cloud\_cover[1]\}}\%.
\end{itemize}

\texttt{<metadata-interpretation-guide>}

\begin{itemize}[leftmargin=1.5em, itemsep=0.25em]
  \item \textbf{Hemisphere:} Positive latitude = Northern; negative = Southern.
  
  \item \textbf{Seasons (month-based):}
  \begin{itemize}[leftmargin=1.5em, itemsep=0.2em]
    \item Northern: Mar-May (spring), Jun-Aug (summer), Sep-Nov (autumn), Dec-Feb (winter).
    \item Southern: Sep-Nov (spring), Dec-Feb (summer), Mar-May (autumn), Jun-Aug (winter).
  \end{itemize}
  
  \item \textbf{Sun azimuth} (sunlight direction, clockwise from north):
  \begin{itemize}[leftmargin=1.5em, itemsep=0.2em]
    \item 0°: north.
    \item 45°: northeast.
    \item 90°: east.
    \item 135°: southeast.
    \item 180°: south.
    \item 225°: southwest.
    \item 270°: west.
    \item 315°: northwest.
  \end{itemize}
  
  \item \textbf{Sun elevation} (above horizon):
  \begin{itemize}[leftmargin=1.5em, itemsep=0.2em]
    \item Low (\(\sim\)0°): long shadows.
    \item Medium (\(\sim\)45°): moderate shadows.
    \item High (\(\sim\)90°): minimal shadows.
  \end{itemize}
\end{itemize}

\texttt{</metadata-interpretation-guide>}

\texttt{<objective>}

Analyze the two satellite images to describe both changes and unchanged aspects between them. Provide a comprehensive analysis that covers visual differences and similarities across all observable elements, including but not limited to key objects, land cover, structures, vegetation, water bodies, urban development, natural features, and any other relevant aspects. Use the provided geospatial metadata only as a reference to inform your analysis, not make it the main focus or describe it directly. Instead, convert relevant metadata into natural, descriptive language without including raw numerical values or technical terms like degrees, percentages, or coordinates. Explain visual influences on changes and unchanged aspects, ensuring descriptions are reasonable, accurate, and not dominated by metadata details or speculation.

\texttt{</objective>}

\texttt{<response-format>}

Output as a structured JSON object:

\begin{lstlisting}[style=jsonstyle]
{
  "temporal_caption": "A comprehensive caption describing both the changes and unchanged aspects between the two images.",
  "Post-temporal_image_generation": "A textual prompt for generating the Post-temporal image, phrased as 'A satellite image showing [detailed predicted changes based on observed differences].'"
}
\end{lstlisting}

\texttt{</response-format>}

\end{tcolorbox}

\captionsetup{type=figure}
\captionof{figure}{Prompt template for draft generation using Qwen3-VL-32B-Instruct in the scalable data construction pipeline.}
\label{fig:draft-generation-prompt}

\begin{tcolorbox}[
  colback=Black!5,
  colframe=Black!70!Black!70,
  title=Text Refinement Prompt,
  breakable,
  enhanced
]
\small

\begin{tcolorbox}[
  colback=Black!5,
  colframe=Black!50!Black!40,
  boxrule=0.3pt,
  left=2pt,right=2pt,top=1pt,bottom=1pt,
  title=System Prompt
]
You are an advanced AI model specialized in refining geospatial analyses, correcting and polishing temporal captions, and enhancing image generation prompts based on satellite imagery insights.
\end{tcolorbox}

You are provided with geospatial metadata and initial analysis outputs for two satellite images of the same geographic location, centered at coordinates (latitude, longitude) (\texttt{\{center\_lat\_lon\}}), but captured at different times. Geospatial metadata includes:

\begin{itemize}[leftmargin=1.5em, itemsep=0.3em]
  \item \textbf{Acquisition times:} Pre-temporal image at \texttt{\{time[0]\}}, Post-temporal image at \texttt{\{time[1]\}}.
  
  \item \textbf{Bounding boxes for key objects} (normalized coordinates in the range [0, 1000)): Pre-temporal image \texttt{\{"bbox\_2d": \{new\_xyxy\_box[0]\}, "label": "\{category[0]\}"\}}, Post-temporal image \texttt{\{"bbox\_2d": \{new\_xyxy\_box[1]\}, "label": "\{category[1]\}"\}}.
  
  \item \textbf{Sun azimuth:} Pre-temporal image \texttt{\{sun\_azimuth[0]\}} degrees, Post-temporal image \texttt{\{sun\_azimuth[1]\}} degrees.
  
  \item \textbf{Sun elevation:} Pre-temporal image \texttt{\{sun\_elevation[0]\}} degrees, Post-temporal image \texttt{\{sun\_elevation[1]\}} degrees.
  
  \item \textbf{Off-nadir angle:} Pre-temporal image \texttt{\{off\_nadir\_angle[0]\}} degrees, Post-temporal image \texttt{\{off\_nadir\_angle[1]\}} degrees.
  
  \item \textbf{Cloud cover:} Pre-temporal image \texttt{\{cloud\_cover[0]\}}\%, Post-temporal image \texttt{\{cloud\_cover[1]\}}\%.
\end{itemize}

The following are the initial outputs from a previous analysis:

\begin{itemize}[leftmargin=1.5em, itemsep=0.3em]
  \item \textbf{temporal\_caption:} \texttt{"\{temporal\_caption\}"}.
  
  \item \textbf{Post-temporal\_image\_generation:} \texttt{"\{post\_temporal\_image\_genera-\\tion\} "}.
\end{itemize}

\texttt{<metadata-interpretation-guide>}

\begin{itemize}[leftmargin=1.5em, itemsep=0.25em]
  \item \textbf{Hemisphere:} Positive latitude = Northern; negative = Southern.
  
  \item \textbf{Seasons (month-based):}
  \begin{itemize}[leftmargin=1.5em, itemsep=0.2em]
    \item Northern: Mar-May (spring), Jun-Aug (summer), Sep-Nov (autumn), Dec-Feb (winter).
    \item Southern: Sep-Nov (spring), Dec-Feb (summer), Mar-May (autumn), Jun-Aug (winter).
  \end{itemize}
  
  \item \textbf{Sun azimuth} (sunlight direction, clockwise from north):
  \begin{itemize}[leftmargin=1.5em, itemsep=0.2em]
    \item 0°: north.
    \item 45°: northeast.
    \item 90°: east.
    \item 135°: southeast.
    \item 180°: south.
    \item 225°: southwest.
    \item 270°: west.
    \item 315°: northwest.
  \end{itemize}
  
  \item \textbf{Sun elevation} (above horizon):
  \begin{itemize}[leftmargin=1.5em, itemsep=0.2em]
    \item Low (\(\sim\)0°): long shadows.
    \item Medium (\(\sim\)45°): moderate shadows.
    \item High (\(\sim\)90°): minimal shadows.
  \end{itemize}
\end{itemize}

\texttt{</metadata-interpretation-guide>}

\texttt{<objective>}

Refine and polish the provided temporal\_caption and Post-temporal\_image\_generation based on the context and metadata guide.

For the temporal\_caption:
\begin{itemize}[leftmargin=1.5em, itemsep=0.25em]
  \item Correct any inaccuracies or inconsistencies with the geospatial metadata.
  \item Improve clarity, conciseness, and flow for better readability.
  \item Ensure it comprehensively covers both changes (e.g., urban development, vegetation shifts) and unchanged aspects (e.g., stable land features) across all observable elements.
  \item Naturally incorporate metadata insights (e.g., lighting effects, seasonal influences) into descriptive language without using raw numerical values or technical terms.
\end{itemize}

For the Post-temporal\_image\_generation:
\begin{itemize}[leftmargin=1.5em, itemsep=0.25em]
  \item Reference key details from the refined temporal\_caption to supplement important information (e.g., add specific predicted changes, visual effects from metadata).
  \item Enhance it to be a more detailed, vivid prompt for image generation, focusing on observed differences and alignments with the caption.
  \item Keep the phrasing as \texttt{`A satellite image showing [detailed predicted changes based on observed differences].'}
  \item Avoid speculation; base enhancements on provided analysis.
\end{itemize}

Use the provided geospatial metadata only as a reference to inform your analysis, not make it the main focus or describe it directly. Instead, convert relevant metadata into natural, descriptive language without including raw numerical values or technical terms like degrees, percentages, or coordinates.

\texttt{</objective>}

\texttt{<response-format>}

Output as a structured JSON object:

\begin{lstlisting}[style=jsonstyle]
{
  "temporal_caption": "Refined comprehensive caption describing both the changes and unchanged aspects between the two images.",
  "Post-temporal_image_generation": "Enhanced textual prompt for generating the Post-temporal image, phrased as 'A satellite image showing [detailed predicted changes based on observed differences].'"
}
\end{lstlisting}

\texttt{</response-format>}

\end{tcolorbox}

\captionsetup{type=figure}
\captionof{figure}{Prompt template for text refinement with Qwen2.5-72B-Instruct in the data construction pipeline.}
\label{fig:text-refinement-prompt}

\begin{tcolorbox}[
  colback=Black!5,
  colframe=Black!70!Black!70,
  title=ST-CQA GPT-Score Prompt,
  breakable,
  enhanced
]
\small

Evaluate a model-generated change caption against a human-generated caption (ground truth) for the SAME area in pre-/post-temporal remote sensing images. Identify the aspects mentioned in the human caption (both changes and unchanged aspects) and calculate the percentage of these aspects correctly mentioned or partially matched in the model caption. Score from 0 to 100, where each aspect contributes equally to the score. Consider similar concepts for partial score. Provide your score (0--100) and a short justification (less than 15 words) in the format of \texttt{`score\#reason'}.\\

Now score the following:\\
Human: \texttt{\{ground\_truth\}}\\
Model: \texttt{\{model\_output\}}\\
Output:

\end{tcolorbox}

\captionsetup{type=figure}
\captionof{figure}{Prompt used by GPT-5-Nano to compute GPT-Score for the spatiotemporal change understanding (ST-CQA) task.}
\label{fig:gpt-5-nano-stcqa-prompt}

\begin{tcolorbox}[
  colback=Black!5,
  colframe=Black!70!Black!70,
  title=RS-WorldModel Stage-1 System Prompt,
  breakable,
  enhanced
]
\small

You are a remote sensing world model. Based on the provided particulars, you can generate future remote sensing imagery at any specified time in the future.

\end{tcolorbox}

\captionsetup{type=figure}
\captionof{figure}{System prompt for RS-WorldModel in Stage 1 (Geo-Aware Generative Pre-training, GAGP).}
\label{fig:rs-worldmodel-stage1-prompt}

\newpage

\begin{tcolorbox}[
  colback=Black!5,
  colframe=Black!70!Black!70,
  title=RS-WorldModel Stage-2/3 System Prompt,
  breakable,
  enhanced
]
\small
You are a remote sensing world model. Based on the provided particulars, you can understand the images and generate future remote sensing imagery at any specified time in the future.

\end{tcolorbox}

\captionsetup{type=figure}
\captionof{figure}{System prompt for RS-WorldModel used in Stage 2 (Synergistic Instruction Tuning) and Stage 3 (Verifiable Reinforcement Optimization).}
\label{fig:rs-worldmodel-stage2-3-prompt}

\begin{tcolorbox}[
  colback=Black!5,
  colframe=Black!70!Black!70,
  title=TFSF GPT-Score Prompt,
  breakable,
  enhanced
]
\small

You are a strict judge for remote sensing future-image prediction.

\texttt{[TEXT\_DESCRIPTION]}\\
\texttt{\{text\_description\}}\\

\texttt{[PROMPT\_METADATA]}\\
\texttt{\{meta\_tse\}}\\

Return TWO integer scores (0--50) and one-sentence reason.

\textbf{(Similarity\_score} (0--50): Text-Image Consistency\\
Evaluate whether the predicted post image matches \texttt{TEXT\_DESCRIPTION}.\\
Use \texttt{PROMPT\_METADATA} only to penalize obvious contradictions (e.g., target cloud=0 but heavy clouds). Do not require exact physics.\\

\textbf{Quality\_score} (0--50): Image Quality\\
Evaluate the predicted post image as a remote sensing image: whether the scene looks realistic for the Earth surface and whether edges/boundaries are visually reasonable.\\
Focus on (1) scene realism/plausibility (no obviously impossible or incoherent land-cover/land-use patterns) and (2) edge/boundary quality (no wobbly/jagged/double edges, halos, boundary bleeding, or melting).\\

Output STRICTLY one-line JSON:\\
\texttt{\{"Similarity\_score": <int 0-50>, "Quality\_score": <int 0-50>, "reason": "<one sentence>"\}}.

\end{tcolorbox}

\captionsetup{type=figure}
\captionof{figure}{Prompt used by GPT-4o to compute GPT-based scores for the text-guided future scene forecasting (TFSF) task.}
\label{fig:gpt-4o-tfsf-prompt}

\newpage

\begin{tcolorbox}[
  colback=Black!5,
  colframe=Black!70!Black!70,
  title=LLM-as-a-Judge Prompt,
  breakable,
  enhanced
]
\small

Evaluate the Model caption against the Human caption (ground truth) for the SAME area in pre-/post-temporal remote sensing images.\\

Compare the key information stated in Human and Model for each dimension (changes, unchanged, time, space, environment). Give full/partial credit for semantically consistent information and reduce for contradictions. Score 0--100 using 5 dimensions (20 points each).\\

Output EXACTLY one line: \texttt{score\#reason} (reason is one sentence).\\

PROMPT\_METADATA:\\
\texttt{\{meta\_tse\}}\\

Human Caption:\\
\texttt{\{ground\_truth\}}\\

Model Caption:\\
\texttt{\{model\_output\}}\\

Output:

\end{tcolorbox}

\captionsetup{type=figure}
\captionof{figure}{LLM-as-a-Judge prompt template based on Qwen3-30B-A3B-Instruct-2507 for verifiable reinforcement optimization (VRO).}
\label{fig:qwen3-llm-judge-prompt}


\end{document}